\documentclass[letterpaper, 10 pt, conference]{ieeeconf}  
\usepackage{amsmath}
\usepackage{graphicx}
\usepackage{color}
\usepackage[obeyspaces]{url}

\usepackage{algorithmic}
\usepackage{algorithm}

\IEEEoverridecommandlockouts                              

\usepackage{subfig}
\title{\LARGE \bf
Configuration-Space Flipper Planning on 3D Terrain
}

\author{Yijun Yuan$^{1}$, Qingwen Xu$^{1}$ and S\"oren Schwertfeger$^{1}$
\thanks{$^{1}$All authors are with the School of Information Science and Technology, 
ShanghaiTech University, China.
        {\tt\small [yuanwj, xuqw, soerensch]@shanghaitech.edu.cn}}%
}

\begin{document}

\setlength{\belowcaptionskip}{-5pt}

\maketitle
\thispagestyle{empty}
\pagestyle{empty}

\begin{abstract}
Flippers are essential components of tracked robot locomotion systems for unstructured terrain, especially within a rescue scenario. Achieving full and semi-autonomy for such rescue robots is the goal of many research efforts. In this work, we propose an algorithm to plan the  morphologies of a small rescue robot with four flippers over 3D ground without any extra sensor, such as pressure sensor. 
To achieve the goal, we simplify the rescue robot as a skeleton on inflated terrain. Its morphology can be represented by configurations of several parameters. Then we plan the mobile movement on 3D terrain with four individually manipulated flippers. We perform real robot experiments on three different obstacles. 
The results show that we move the flippers very effectively and are thus able to tackle those terrains very well.
\end{abstract}

\section{Introduction}
\begin{figure}[tpb]
	\centering
	\subfloat[Robot]{
		\label{fig:robot:robot}
		\includegraphics[width=0.5\linewidth]{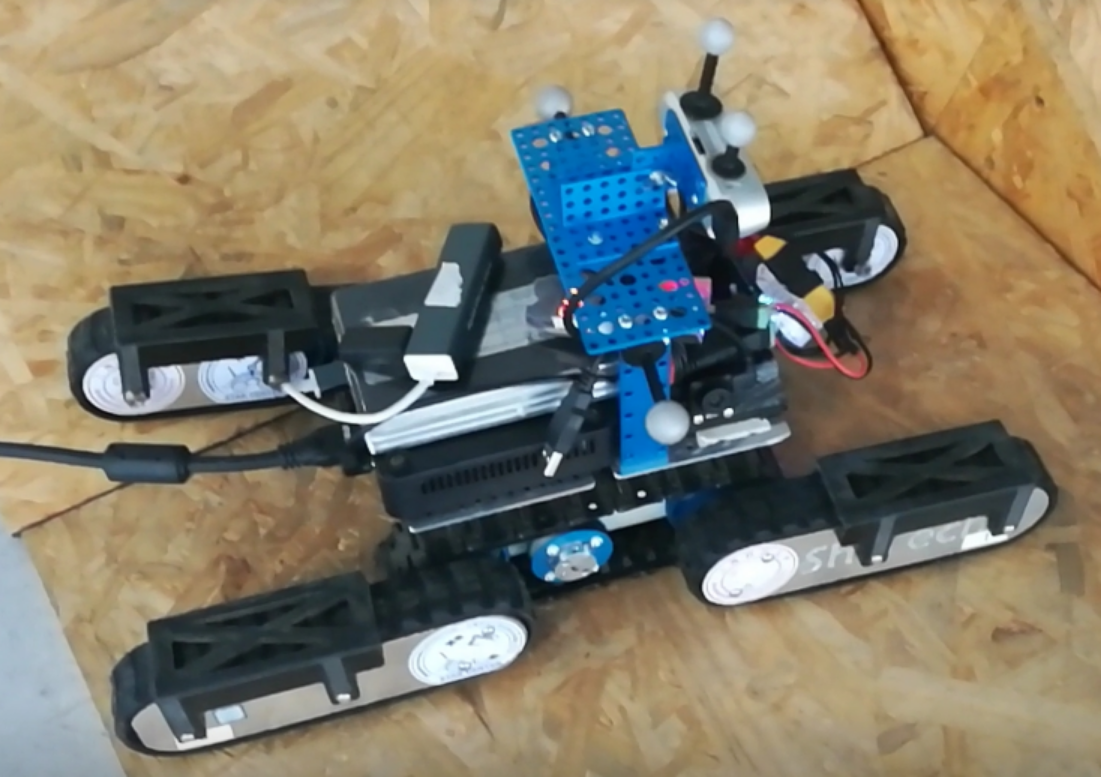}}
	\subfloat[Model]{
		\label{fig:robot:model}
		\includegraphics[width=.5\linewidth]{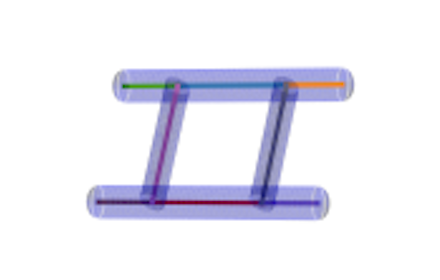}}
	\caption{Rescue Robot and its simplified model.}
	\label{fig:robot}
\end{figure}

Recently, rescue robots are expected to play an important role in search and rescue after disasters happen \cite{sheh201616} or in the military \cite{choi2019development}. Some rescue robots are designed like animals, such as snakes \cite{miller200213,konyo2008ciliary} and dogs \cite{ferworn2012dog}, to increase mobility, which is mechanically complex. An easy alternative approach is to use tracked robots for propulsion. However, tracked robots usually meet the problem that big obstacles are higher than the height of the robot base. To increase the mobility of tracked robots, sub-tracks or flippers are added to the propulsion system, as shown in Fig.~\ref{fig:robot:robot}. Moreover, those flippers are even more important for small rescue robots compared to big ones, since they strongly rely on their flipper when traversing through rough terrain. 

Even though the tracked robots with flippers have good mobility, they have a high cognitive load on the operators, due to their many degrees of freedom, as mentioned in \cite{nagatani2008semi,okada2011shared}. Therefore, autonomous flipper planning is really helpful for operators and autonomous path planning is important for rescue robots. This then also leads the way towards full autonomy in rescue robotics \cite{pathak2007fully}. Some researchers designed flipper behaviors based on the experience of expert operators. The motion strategy used in \cite{okada2011shared} is based on the operation by skilled operators. Sheh et al. proposed behavioral cloning from human experts when the robots traversing on the rugged terrains \cite{sheh2011behavioural}. 

In addition to experience from operators, some work proposed autonomous or semi-autonomous control of the rescue robots with flippers based on its kinematics and physical model. To design a close-loop control system, the rescue robots should "know" the environment and their own states. To build such a system, Ohno et al. added encoder, current and gravity sensors to measure the state of the robot so that the robot can adjust its pose on the basis of data from these sensors \cite{ohno2007semi}. In their follow-up work \cite{rohmer2010integration}, they also mounted laser scanners to provide the environment sensing and automatically adjusted the flippers according to the environment. Moreover, Pecka et al. even utilized a robot arm to gather data which cannot been collected directly by the robot base \cite{pecka2016controlling}.

\begin{figure*}[tb]
	\centering
	\includegraphics[width=0.98\linewidth]{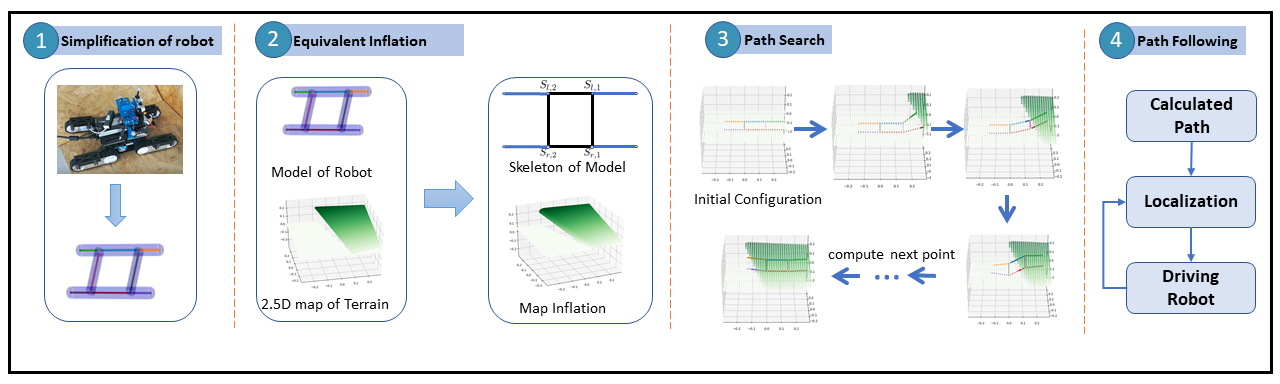}
	\caption{System Overview}
	\label{fig:pipeline}
\end{figure*}

No matter if the environmental data is gathered by the robots' sensors in the real environment or if it is given by system settings in a simulation platform, the execution performance for rescue robots traversing on rugged terrain is always one of the biggest challenge, due to gravity and disturbance. To make the robot follow the given path, Martens et al. designed a feedback system to compensate the asymmetric wheels and gravity effect for a mobile robot to climb stairs \cite{martens1994stabilization}. Inspired by this work, Steplight et al. extended the remote-control strategy to autonomous stair-climbing with additional sensors to detect the environment features \cite{steplight2000mode}. Additionally, a pre-defined morphology is performed during climbing. In \cite{colas20133d}, four different driving modes of the robot base are used to divide climbing into four periods and flippers are adjusted to meet the orientation requirement. 
Zimmermann et al. defined five flipper modes which are applied in different period of traversing complex terrain \cite{zimmermann2014adaptive}. In addition, to make sure that the robot is stable on rugged terrains, most control strategies take the center of mass and gravity into consideration. For example, Vu et al. analyzed  the moment balance and stability of a customized robot when it is on the stairs with different pose \cite{vu2008autonomous}. In \cite{gianni2016adaptive}, stability is interpreted as each flipper and robot base should touch the ground.

Besides, there are data-based methods that can train a mapping from state to action \cite{zimmermann2014adaptive, pecka2016autonomous, sokolov2017hyperneat}, which can learn to use the flippers. But the weakness is the limited coverage of the training terrain, which in turn might cause a crash due to the overfitted parameters. 


In the above studies, autonomous flipper planning is based on 2D planning, where the front two flippers are under the same motion and the rear two are the same, such as \cite{rohmer2010integration, zimmermann2014adaptive, pecka2016controlling,yuan2019configuration}. Only when the flipper is adjusted slightly to touch the ground, the left and right flippers may execute to different angles \cite{ okada2011shared, gianni2016adaptive}.  In this work, we propose an active control method for flipper planning, which calculates the safe morphology and feasible path for the robot with flippers in 3D space. Related to our previous work \cite{yuan2019configuration}, that made the simplification of flippers in 2D space, we consider the four flippers individually in  the 3D terrain, which greatly increases the mobility of the robot. In our algorithm we maintain the analysis in continuous configuration space when calculating the morphology of the robot.

The contributions of this work are summarized as follows:
\begin{itemize}
	\item Continuous space flipper planning in 3D terrain for small sub-tracked robot.
	\item A real robot implementation to follow the path of configuration with thorough experimental evaluation.
\end{itemize}

The rest of the paper is structured as follows: The important modules of the approach are presented in Section~\ref{sec::method}. In Section~\ref{sec:exp} we introduce the experiment settings and discuss the experimental results. Finally, we draw conclusions in Section~\ref{sec:conclusion}. 

\section{Methodology}
\label{sec::method}
\begin{figure}[tpb]
	\centering
	\includegraphics[width=0.83\linewidth]{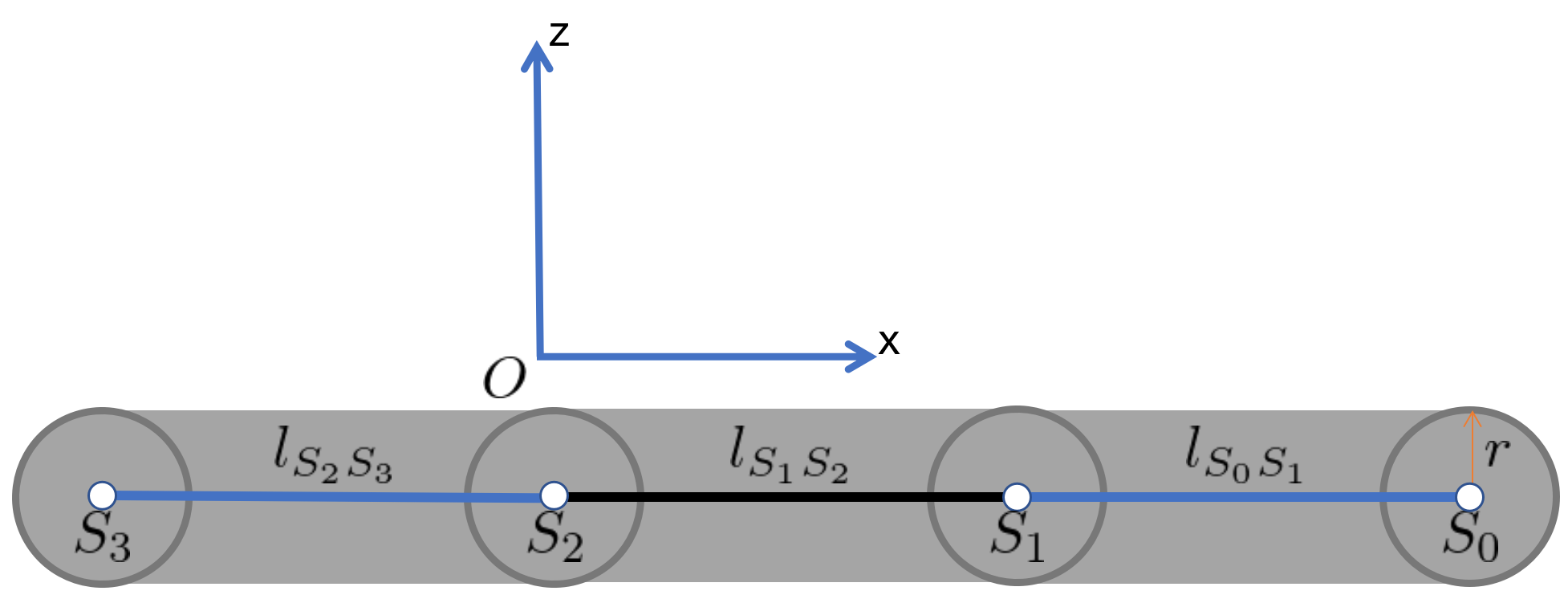}
	\includegraphics[width=0.83\linewidth]{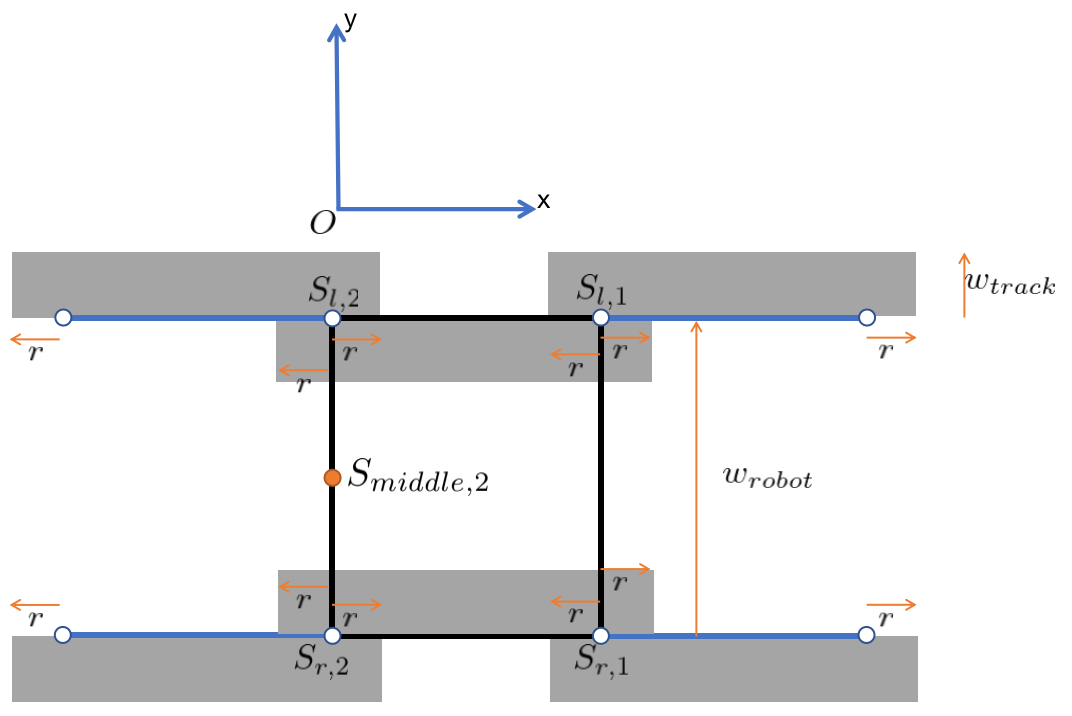}	
	\caption{Side-view and top-view sketch for Fig.~\ref{fig:robot:robot}. The skeleton is drawn with black and blue lines connecting the white joints. The robot's coordinate system's origin $O$ is $S_{middle, 2}$, with the $x$-axis from back to front (left to right in the image), the $y$-axis to the left of the robot and the $z$-axis pointing up.}
	\label{fig:sketch}
\end{figure}
To make it easier to analyze the robot morphology on the terrain, similar to what we did previously in \cite{yuan2019configuration}, we transform the robot and map to skeleton and inflated map. We generate the skeleton-inflated ground representation in 3D space and then compute the configuration. After that, a customized path search is correspondingly utilized to find a sequence of morphologies for the robot.

We work with several simplifying assumptions:
\begin{itemize}
	\item the center of gravity is always in the center of the robot base;
	\item no slip and floating;
	\item driving forward with a steady speed.
\end{itemize}

\subsection{Overview}
\label{sec:overview}
The overview of this system is shown in Fig. \ref{fig:pipeline}. 
Given a 2.5D map of the environment and the robot parameters, the workflow of this approach is as follows:
\begin{itemize}
	\item Simplify the robot as model as in Fig. \ref{fig:robot},
	\item Equivalently morph the robot model and 2.5D map to the robot skeleton and inflated map representation as in Section~\ref{sec::transform},
	\item Build a path of configurations to the goal, see Section~\ref{sec::path}. The function to calls to generate the robot pose and flipper parameters are in Section~\ref{sec::generation}.
	\item Given a path, to follow the trajectory of configurations for a real robot implementation, shown in Section~\ref{sec::following}.
\end{itemize}

\subsection{Robot Simplification}
\label{sec::Simplify}

Following \cite{yuan2019configuration}, a simple robot model is needed to make 3D flipper planning feasible.


In Fig. \ref{fig:sketch}, we can find the top and side view sketch of the robot in Fig. \ref{fig:robot:robot}. For this robot in Fig. \ref{fig:robot:robot}, the values are $r=3.5cm$, $w_{track}=3cm$. This sketch describes the robot with points $S_{l/r,0/1/2/3}$ and lines connecting those points. 

Around those blue and black lines in Fig. \ref{fig:sketch}, we generate a surface that the closest distance from blue lines to each surface point is $r$. The simplified model is as Fig. \ref{fig:robot:model}. 

\subsection{Equivalent Inflation}
\label{sec::transform}
To compute the morphology of the robot, simplification can bring about the convenience. In this paper, our simplification also consist of two parts: representing robot as a skeleton and inflating the ground.

\subsubsection{Skeleton Representation}
\label{simplify}

\begin{figure}[b]
	\centering
	\includegraphics[width=0.63\linewidth]{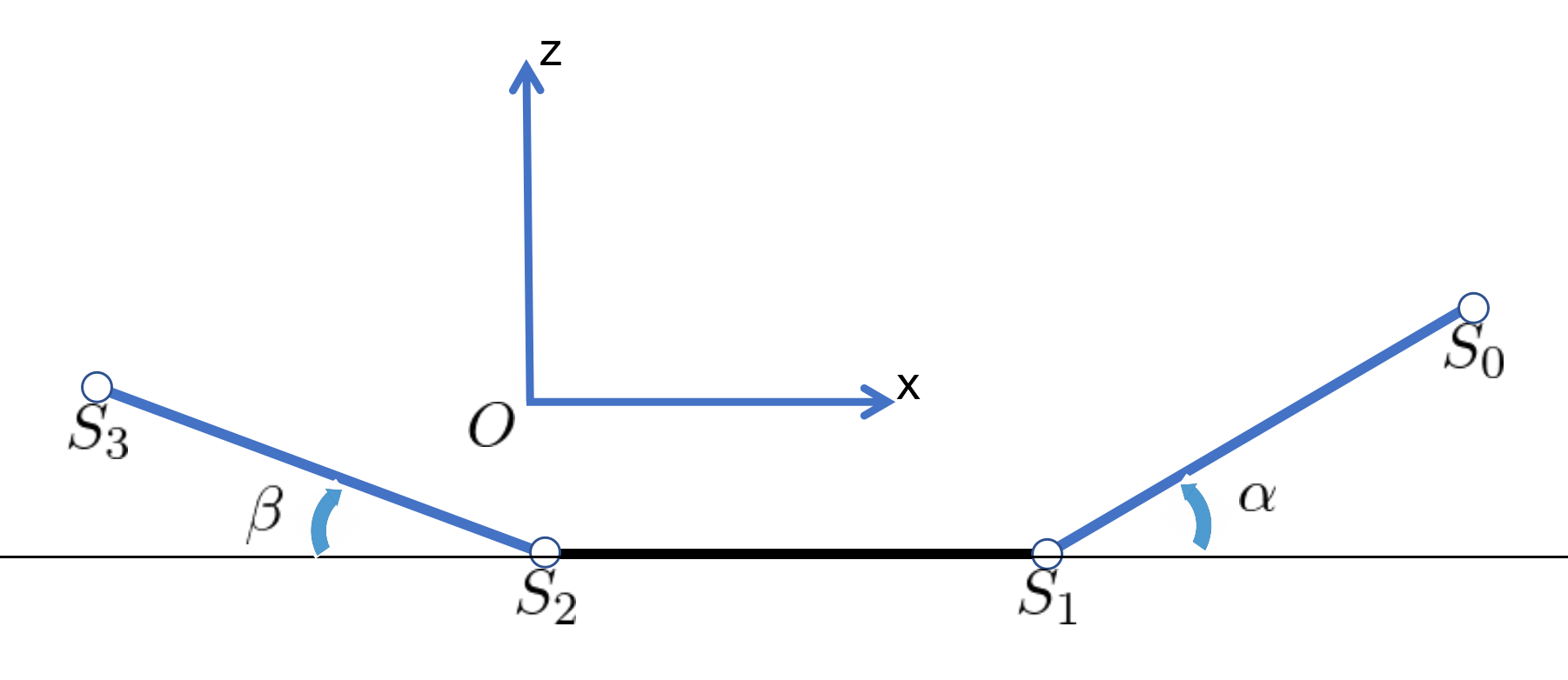}
	\caption{The flipper angles.}
	\label{fig:angle}
\end{figure}

In Fig. \ref{fig:sketch}, we can find the top view and side view sketch of robot in Fig. \ref{fig:robot}. 

The model simplification is on top of the tracked rescue robot with four sub-tracks, which use black and blue lines as the skeleton as in Fig.~\ref{fig:sketch}. We use $S$ to denote a joint on the skeleton. $S_l$ is on the left side and $S_r$ is on the right side. 

Initially, as in Fig.~\ref{fig:sketch}, the robot is facing to the positive $x$ axis, the $y$ axis is to the robot left, the $z$ axis is pointing up. The roll, pitch, yaw angles are around $x$, $y$ and $z$, respectively, following the right hand rule.

In our implementation the order of the three Euler angles is in yaw ($\psi$) , pitch ($\theta$), roll ($\phi$). In this paper, our robot is set to always move forward, thus $\psi$ is always set to $0$.

Also following \cite{yuan2019configuration}, we use $\alpha$ and $\beta$ to denote the angle of the front and back flippers, as in Fig. \ref{fig:angle}, with a subscript $l$ or $r$ to specify the side.

\subsubsection{Inflated Map}
\label{sec::distanceMap}
Since the robot is now represented as a skeleton, the map should be inflated accordingly.

As our rescue robot is moving on the ground, we consider it is adequate to use a 2.5D elevation map to represent the ground scene. We are thus able to do the inflation on that map with the proposed method.

Given a 2.5D map, it consists of $(x,y)$ point $p_i\in \Re^2$, with $i \in {1,2,3,\cdots,N}$, where $N$ is the number of ground points in the grid map.
Then the value on the map that is the height $h(p_i)$.

Correspondingly, since we simplified the robot to a skeleton, the inflated ground map has to ensure that the closest distance from skeleton to map is no less than wheel radius.

Following \cite{yuan2018incremental}, we build the distance map on the original ground with a special kernel $H_{r,h}$. Given a kernel with height $h$, the point $p_j$ with $(\Delta x, \Delta y)$ to the kernel center has value:
\begin{multline}
H_{r,h}((\Delta x, \Delta y))=  \\
\begin{cases}
h+\sqrt{r^2-\Delta x^2 - \Delta y^2}, & r^2 \leq \Delta x^2 + \Delta y^2,\\
0, & \text{otherwise}.
\end{cases}
\end{multline}


We represent a ground pixel on position $p_i$ as a delta function $\delta(x-p_i)$. The function with input location $q$ generates a distance map of each point $i$ as a convolution of $\delta(x-p_i)$ and a kernel $H_{r,h(p_i)}$ is:
\begin{equation}
	D_i(q) = H_{r,h(p_i)}\ast \delta(q-p_i)
\end{equation}

Then the function to generate the inflated map can be represented as: 
\begin{equation}
D(q) = \max_{i=1}^ND_i(q)
\end{equation}

\begin{figure}[]
	\centering
	\subfloat[Height Map]{
		\includegraphics[width=.5\linewidth]{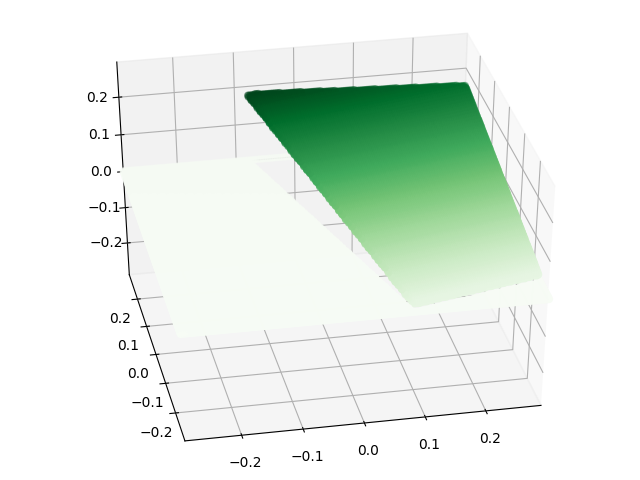}		
	}
	\subfloat[Inflated Map]{
		\includegraphics[width=.5\linewidth]{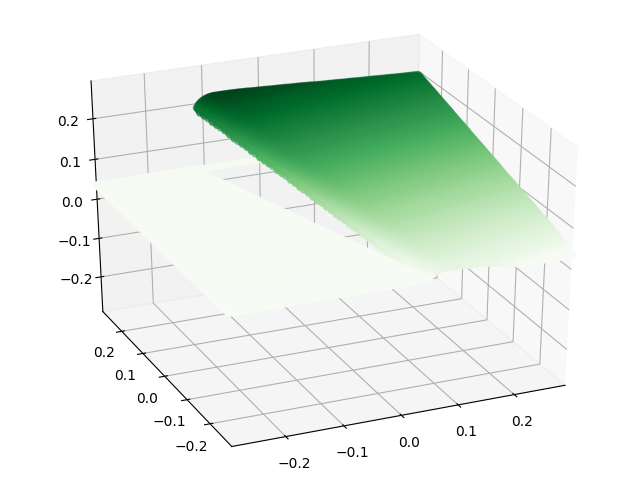}		
	}	
	\caption{Height map and its inflated map with $r=0.035m$.}
	\label{fig:dilated}
\end{figure}

One example result is shown in Fig. \ref{fig:dilated}, where one possible obstacle is found.

\subsection{Configuration Generation}
\label{sec::generation}
For the path search in Section~\ref{sec::path}, the rescue robot is considered as a mass point. This subsection of Configuration Generation is the function utilized to provide that abstraction.
The goal of the configuration generation is to find a batch of possible morphologies at certain position.

In this part, each configuration is designed from the skeleton-inflated ground and can be transformed correspondingly to a robot-ground scene.

The configuration consists of three parts: \textbf{a.} the Euclidean position, \textbf{b.} the orientation from the pose and \textbf{c.} the flipper angles. 

Given the location $x,y,z$ of the reference point, this configuration generation function will: \textbf{(1)} generate a range of possible orientations for the robot base, see Algorithm \ref{alg:col_param}. Also, for each pair of location and orientation, it can \textbf{(2)} uniquely determine the four flipper angles via collision checking, see Algorithm \ref{alg:exp_param}.

During the path search, a sequence of possible heights $z$ are provided for each $x,y$ location.

The orientation can be described with the yaw, pitch, roll ($\psi$, $\theta$ and $\phi$). It is not convenient to search possible orientation by setting the center of robot as the reference point, when we want to compute the possible morphologies (especially rotation angle) on top of it. 
Thus, following \cite{yuan2019configuration}, $S_2$ has been utilized as the reference point. Because we have $S_{l,2}$ and $S_{r,2}$ on the left and right side, when doing path search in next Section~\ref{sec::path}, both sides will be utilized to collect the candidates. 
The flipper angles can be deterministically determined given the position and orientation.

To ensure the safety of the robot, we constrain that for both sides of line $S_2 S_1$, at least one point is touching the ground. The implementation is done by searching over the pitch and roll to find the possible candidates. 

\begin{algorithm}[htbp!]
	\small
	\caption{Get orientation parameters. \small{(Assume left reference point.)}}
	\label{alg:col_param}
	\begin{algorithmic}[1]
		\STATE \textbf{Input:} Left reference point xyz location $P_{S_{l2}}$; inflated map D.
		\STATE	$\theta s$ $\leftarrow$ pitch candidates for line $S_{l2}S_{l1}$ with $\psi=0$ and $\theta \in [ \theta_{lb}, \theta_{ub} ]$
		\FORALL{ $\theta \in \theta s $} 
		\STATE $\phi$ $\leftarrow$ get roll candidate(fix $S_{l2}S_{l1}$ as axis, make line $S_{r2}S_{r1}$ touch map surface D.)
		\STATE Add $(P_{S_{l2}},\psi, \theta, \phi)$ to pose parameter set $\mathbf T$.
		\ENDFOR
		\STATE \textbf{Output:} $\mathbf T$
	\end{algorithmic}
\end{algorithm}

Assume we use the $S_{l2}$ as reference point and $P$ is its 3D location. The $\psi$ is fixed in the very beginning as $0$. Then we can rotate $S_{l2}S_{l1}$ around the $y$ axis with $S_{l2}$ as the rotation center. 

If $S_{l2}$ is not on the ground, we should find the smallest $\theta$, the upper bound $\theta_{ub}$ that makes $S_{l2}S_{l1}$ touch the ground.

If $S_{l2}$ is on the ground, we can also correspondingly find the smallest $\theta$, the lower bound $\theta_{lb}$ that make line $S_{l2}S_{l1}S_{l0}$ touch the ground. And the chosen $\theta \in \{ \theta_{lb}, \theta_{ub}  \}$.

Then with each solved $S_{l2}S_{l1}$, we can compute the roll $\phi$ by finding the smallest angle that makes the rectangular plane $S_{l2}S_{l1}S_{r1}S_{r2}$ touch the ground.

So now position and orientation have been solved, the only things left are the flipper angles to support such a pose. Now we collect a batch of pose candidates as in Algorithm \ref{alg:col_param}.

\begin{algorithm}[htbp!]
	\small
	\caption{Get flipper parameters. \small{(Assume left reference point.)}}
	\label{alg:exp_param}
	\begin{algorithmic}[1]
		\STATE \textbf{Input:} $P_{S_{l2}},\psi, \theta, \phi$; inflated map D
		\STATE get the location of points $S_{r2}, S_{l1}, S_{r1}$.
		\STATE $\alpha_l$ $\leftarrow$ get angle by rotate $S_{l1}S_{l0}$ around axis $S_{r1}S_{l1}$ to make  line $S_{l1}S_{l0}$ colliding map surface D.
		\STATE Similarly compute the $\alpha_r$, $\beta_l$, $\beta_r$.
		\STATE \textbf{Output:} $(\alpha_l, \alpha_r, \beta_l,\beta_r )$
	\end{algorithmic}
\end{algorithm}

Given the fixed joint (obtained from fixed location and orientation) on each flipper, we can uniquely get the flipper setting by finding the angle that make flipper touch ground surface (not puncture the surface). 


\subsection{Path Search}
\label{sec::path}
In \cite{yuan2019configuration}, to generate a path from source to goal, a configuration space is generated with robot as a mass point. Then it applies a greedy search to find a path. 

However, in our 3D terrain, with 6 degree pose plus 4 degree flipper parameter, it is inappropriate to build such a big configuration space covering source to goal.
Thus, instead, we search for a next path point from a configuration space of next step given the current point. They are equivalent because during the greedy search, only a local region that are close to current point should be considered as the candidates of next step.

Given configuration of current point
, $S_{l,2}$ and $S_{r,2}$ can be extracted. Use both left side and right side $S_2$ as reference point respectively, we get the incrementally moved $x, y$ for next step, then we samples a sequence of possible height $z$ based on current height of reference point. Then for both left and right reference point of next step
it calls function in Algorithm. \ref{alg:col_param} to get batches of possible orientation. 

Next we use a cost function to evaluate the safety on gravity and extract best orientation, $z$. 

After that, Call function in Algorithm. \ref{alg:exp_param} to compute the flippers angles.

Since we fixed the yaw $\psi$ as 0, the $x$ of the reference point will always increase.
The goal of this part is to make robot move forward to cross a certain terrain. 

\begin{algorithm}[htbp!]
	\small
	\caption{Path Search.}
	\label{alg:path_search}
	\begin{algorithmic}[1]
		\STATE \textbf{Input:} initial $P_{S_{l,2}}, P_{S_{r,2}},\psi, \theta, \phi$; inflated map D; $P_{target}$; $\Delta x$; $\Delta h$
		\STATE 
		\WHILE{not reach target}
		\STATE $P_{ref}\leftarrow P_{S_{l,2}}$
		\STATE $P_{ref}.x += \Delta x$
		\FORALL{$dh \in \{0,\Delta h, 2\Delta h, \cdots\}$}
			\STATE $P_{ref}.z = h(P_{ref}.x,P_{ref}.y)+dh$
			\STATE $T_l\leftarrow getPose(P_{ref},D)$ \COMMENT{Algorithm. \ref{alg:col_param}}
		\ENDFOR
		\FORALL{pose in $T_l$}
			\STATE Add cost($pose$) into the costs set $C$
		\ENDFOR
		\STATE Similarly get $T_r$ with right side reference point and expand $C$.
		\STATE $P_{S_{l,2}}, P_{S_{r,2}},\psi, \theta, \phi \leftarrow$ with smallest cost in $C$ over $T_l$ and $T_r$
		\STATE $\alpha_l, \alpha_r, \beta_l, \beta_r \leftarrow getFlipper(P_{S_{l,2}}/P_{S_{r,2}},D)$ \COMMENT{Algorithm. \ref{alg:exp_param}}
		\STATE Add $(P_{S_{l,2}}, P_{S_{r,2}},\psi, \theta, \phi,\alpha_l, \alpha_r, \beta_l, \beta_r)$ into $path$
		\ENDWHILE
		\STATE \textbf{Output:} $path$
	\end{algorithmic}
\end{algorithm}

Given morphology \[M_i=(P_{S_{l,2}}, P_{S_{r,2}},\psi, \theta, \phi,\alpha_l, \alpha_r, \beta_l, \beta_r)\] at frame $i$, we use both $S_{l2}$ and $S_{r2}$ to compute its pose in the next step and collect a batch of possible candidates.

For each reference point $S_{2}$, we let the reference point move forward a step on the $x$ axis, sample a sequence of heights of the reference point and compute its orientation and flipper angle as in Algorithm \ref{alg:path_search}. 

Then we use cost function to find the most gravity safe option. The cost function is the sum square of difference between the inflated map height $h((P.x,P.y))$ and the point height $P.z$ with $P$ that is in the middle line of robot base. Because for inflated map, on the same location, smaller the height difference in points from middle line indicates it is closer to the ground while less slope on the robot base.

We show one possible path in Fig. \ref{fig:path} to demonstrate the morphology at each configuration point.

\newcommand\pathlen{0.25}
\begin{figure*}[htbp!]
	\centering
	\includegraphics[width=\pathlen\linewidth]{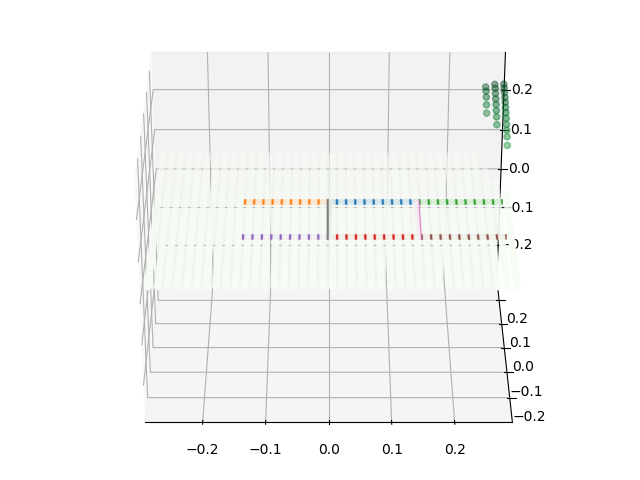} \hspace{-1em}
	\includegraphics[width=\pathlen\linewidth]{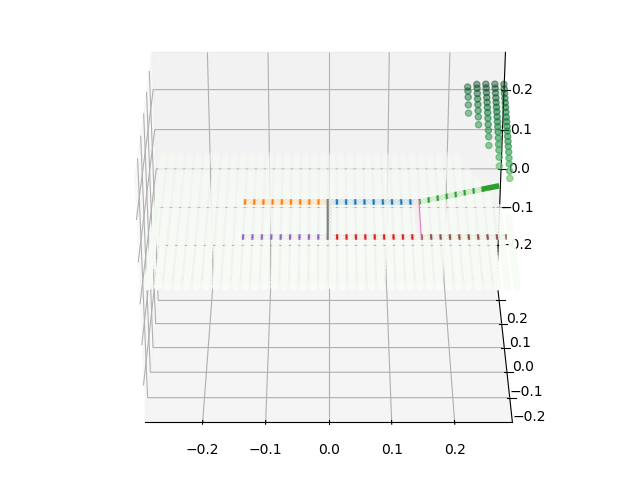}\hspace{-1em}
	\includegraphics[width=\pathlen\linewidth]{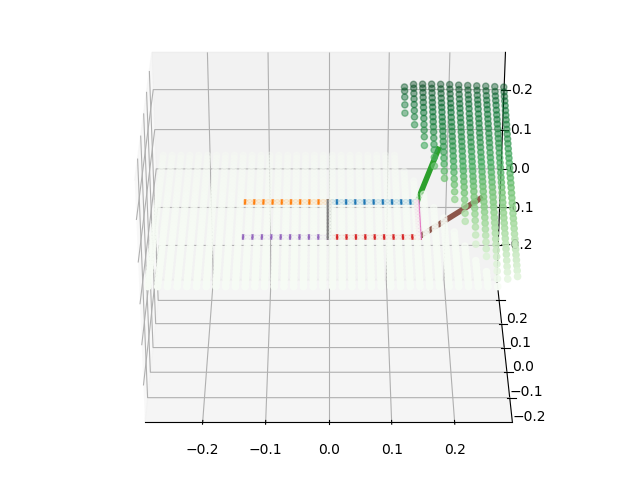} \hspace{-1em}
	\includegraphics[width=\pathlen\linewidth]{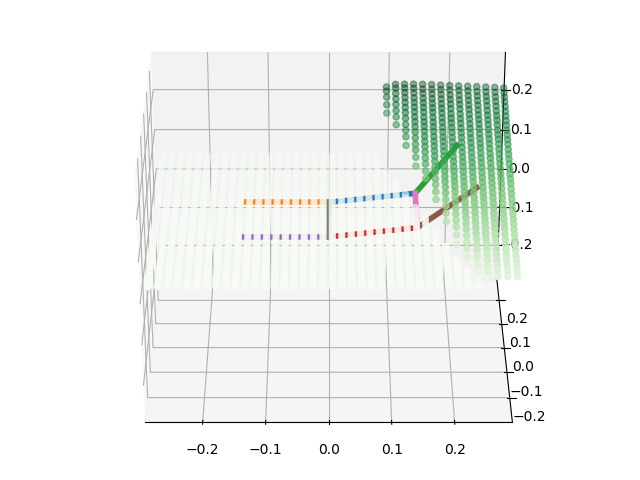} \\
	\includegraphics[width=\pathlen\linewidth]{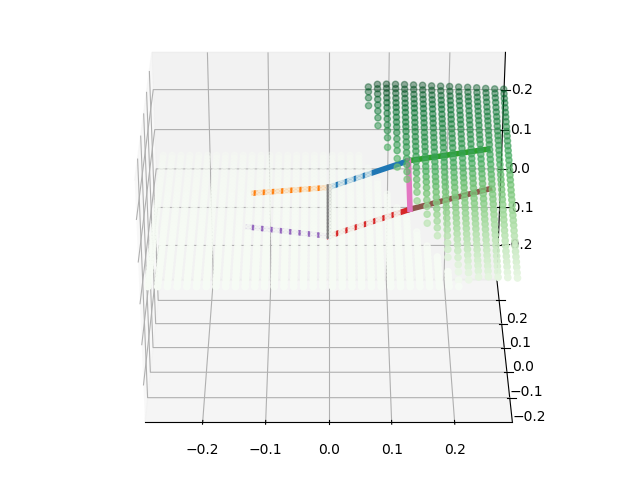} \hspace{-1em}
	\includegraphics[width=\pathlen\linewidth]{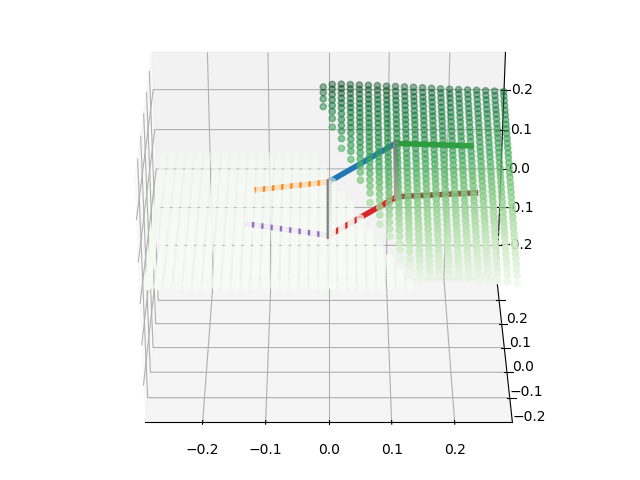} \hspace{-1em}
	\includegraphics[width=\pathlen\linewidth]{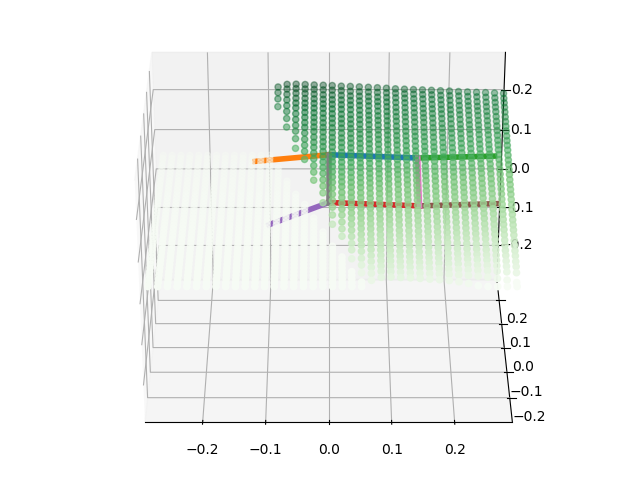} \hspace{-1em}
	\includegraphics[width=\pathlen\linewidth]{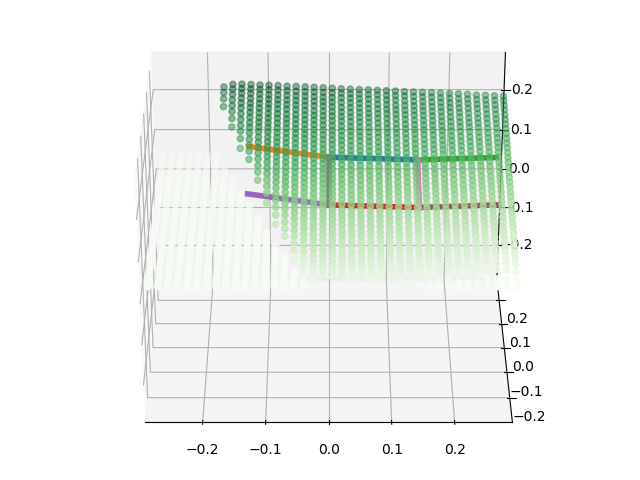}
	\caption{A possible path to get over the iramp with $15^{\circ}$ rotation in experiment. Each subfigure is a configuration point on the path.}
	\label{fig:path}
\end{figure*}

\subsection{Path Following}
\label{sec::following}
The path following can be divided into a sequence of subproblems that robot moves from one point to the next.

Given $M_i$ and $M_{i+1}$, we need to compute the track movement to make robot follow.

Different from \cite{yuan2019configuration}, that can compute the distance to make the track move accurately, in this 3D scenario, it becomes much more complicated. Thus real time localization of the robot in map is required to let robot follow the planned path.

To realize the goal, while the robot is driving, we make sure that the robot keeps the yaw $\psi$ unchanged by speed up one side a little to fix the bias. We check if the robot reached the target by tracking point $S_{middle,2}$ (middle point between $S_{l,2}$ and $S_{r,2}$).  


\begin{figure}[t!]
	\centering
	\subfloat[step]{
		\includegraphics[width=.32\linewidth]{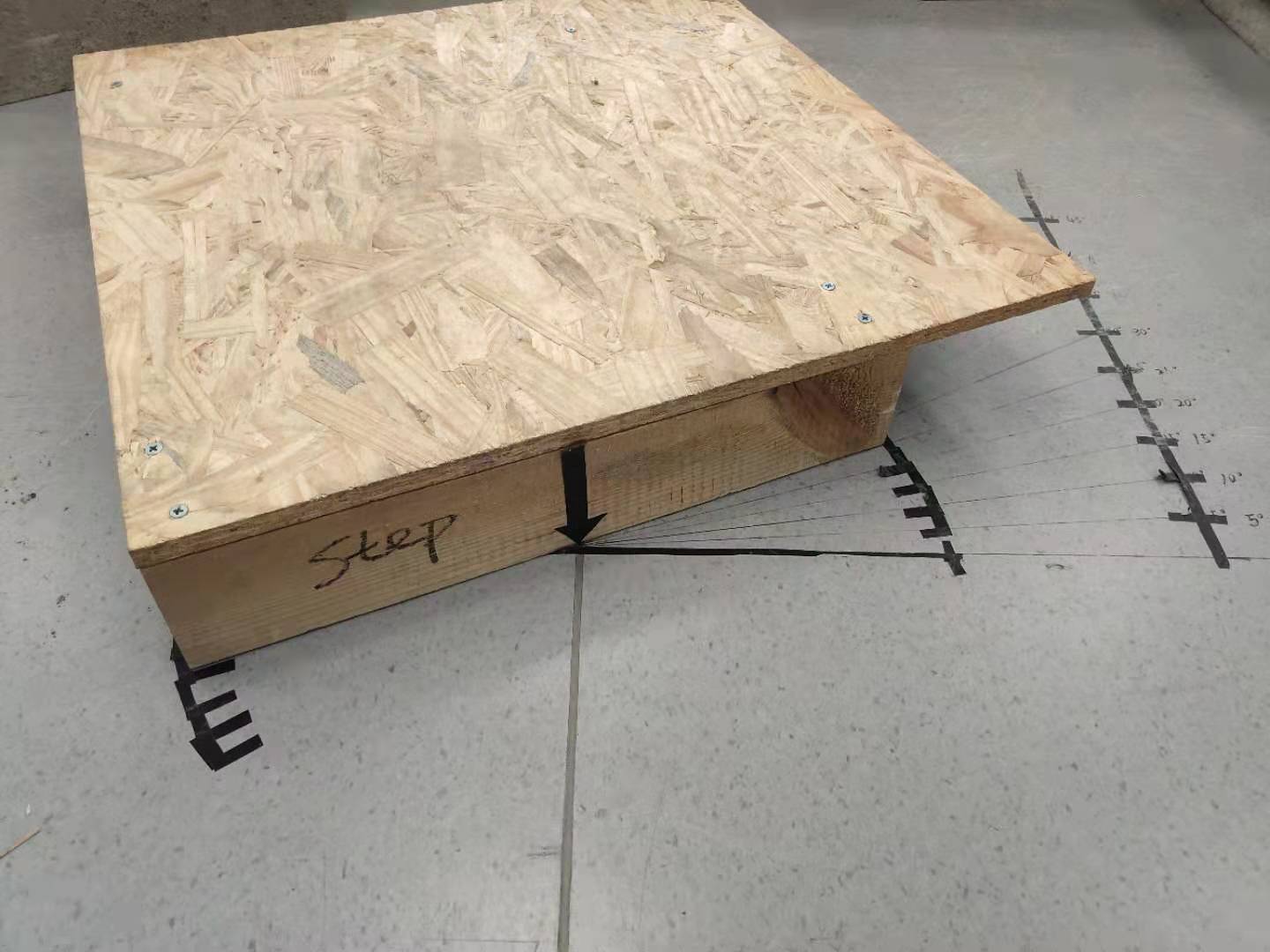}		
	}
	\subfloat[ramp]{
		\includegraphics[width=.32\linewidth]{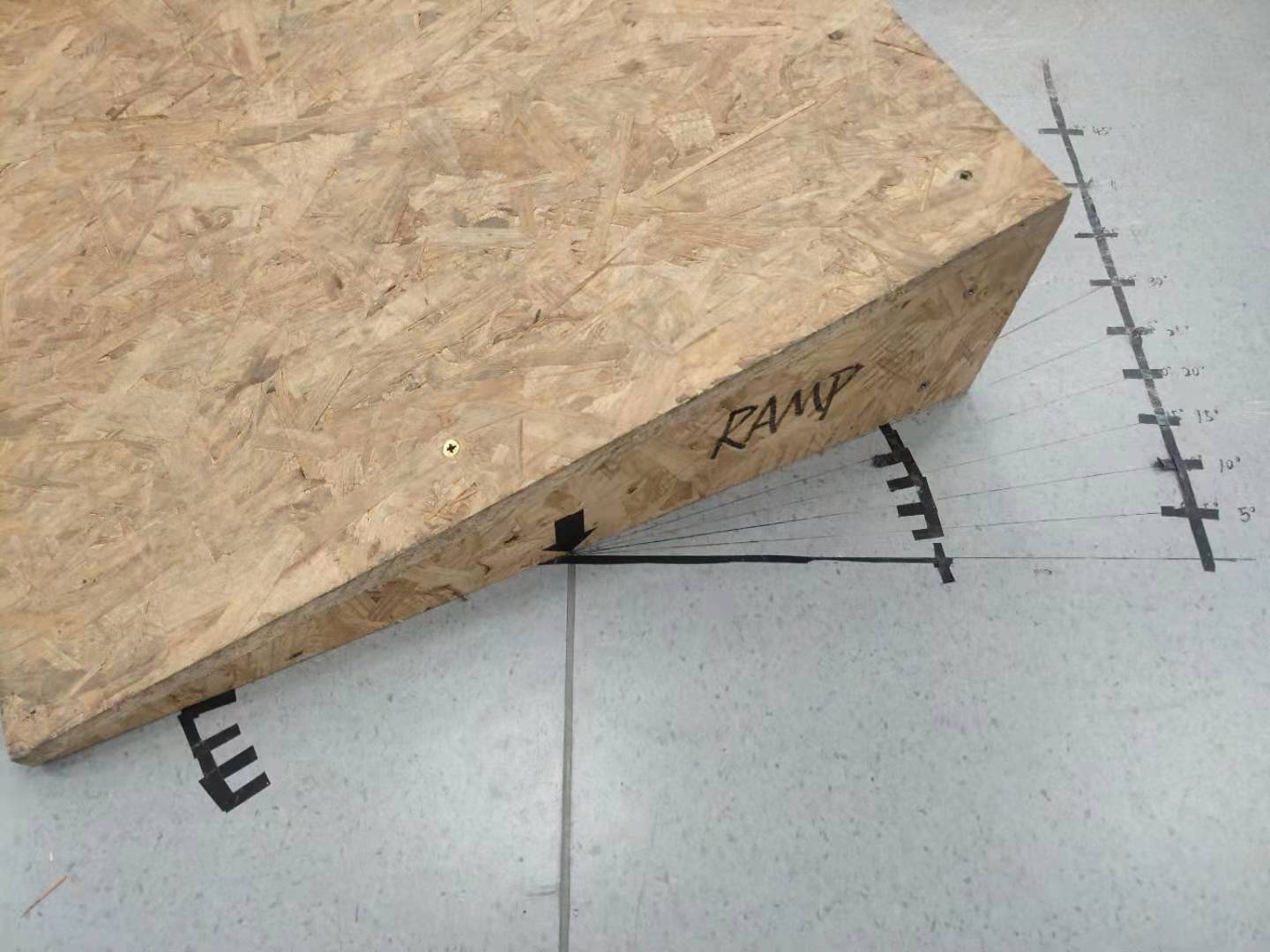}
	}
	\subfloat[iramp]{
		\includegraphics[width=.32\linewidth]{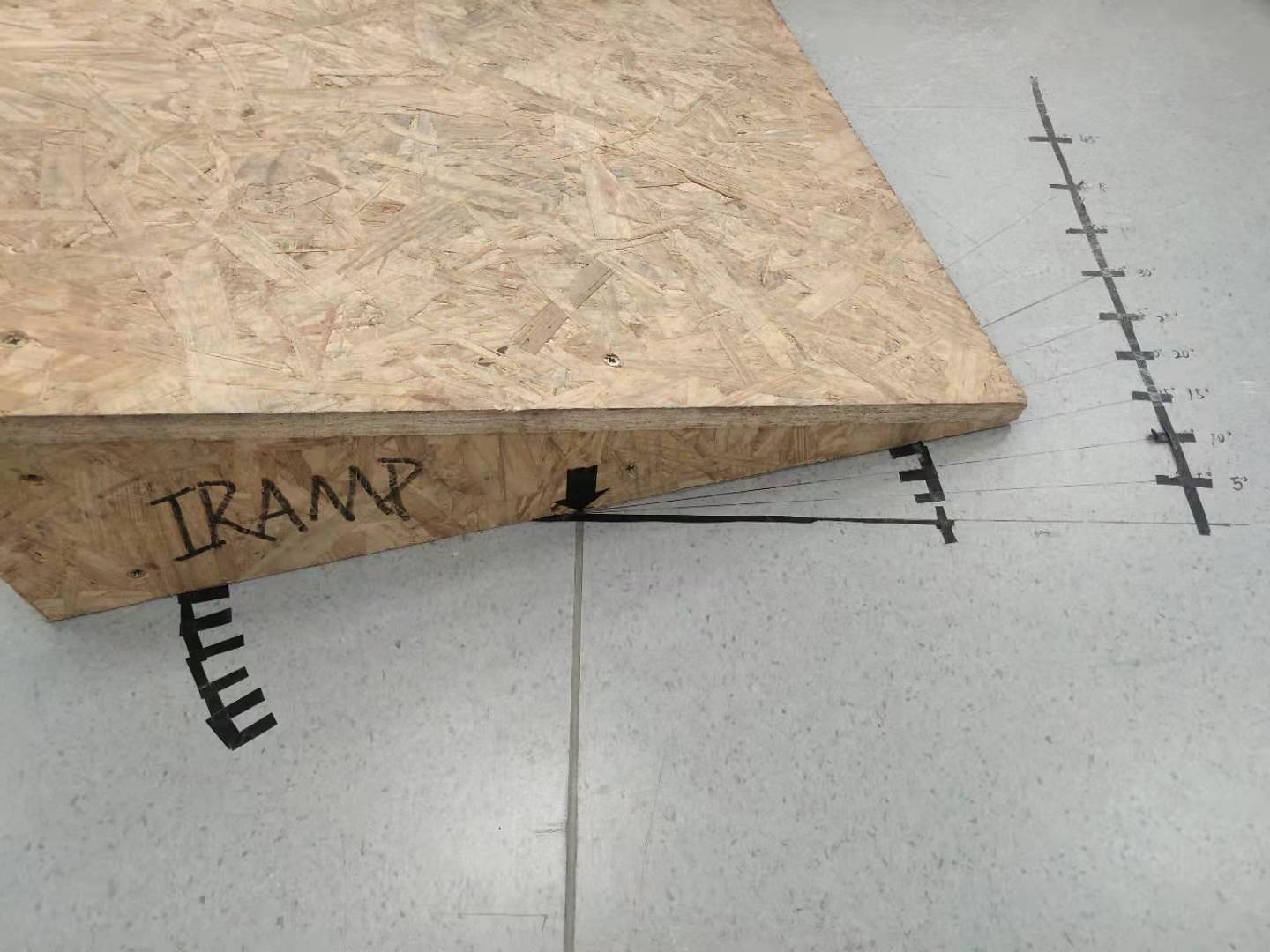}
	}
	
	\caption{The obstacles in the experiments. The robot is orientated to the rotation axis of obstacle with same distance from the $S_{middle,2}$.}
	\label{fig:obstacle}
\end{figure}

\newcommand\locSize{0.12}
\begin{figure*}[t!]
	\centering
	\subfloat[step]{
		\label{fig:simple_bias_loc:step}
		\includegraphics[width=\locSize\linewidth]{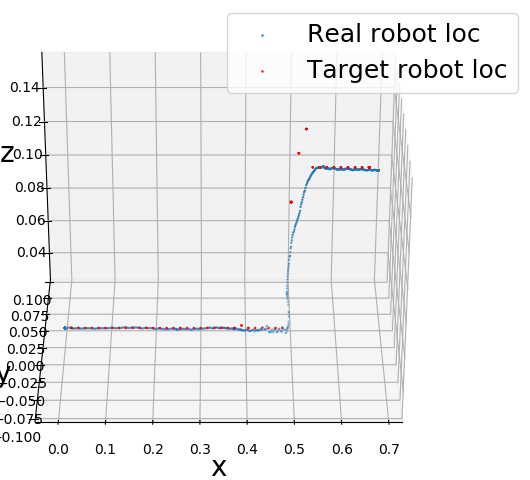}
		\includegraphics[width=\locSize\linewidth]{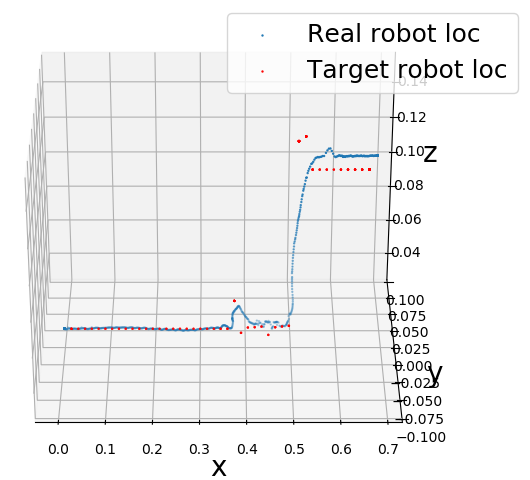}
		\includegraphics[width=\locSize\linewidth]{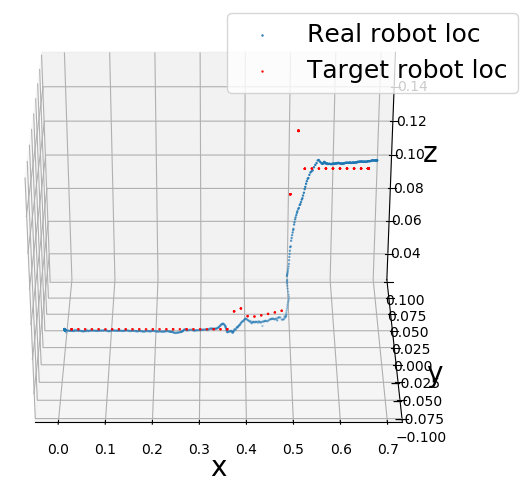}
		\includegraphics[width=\locSize\linewidth]{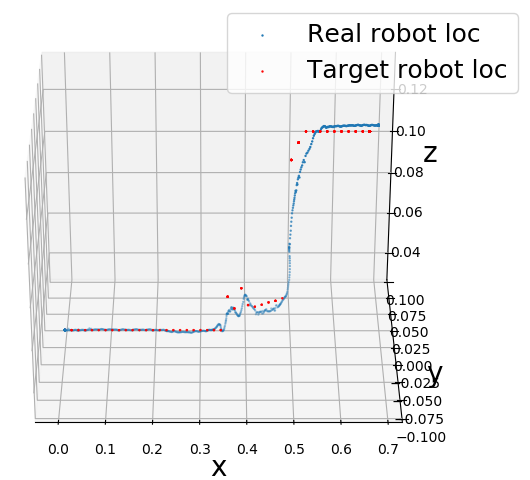}
		\includegraphics[width=\locSize\linewidth]{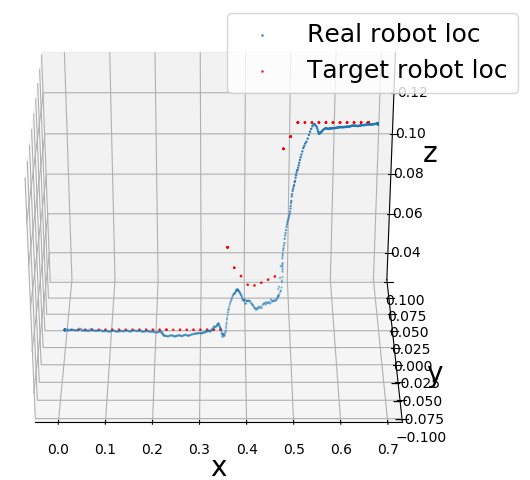}
		\includegraphics[width=\locSize\linewidth]{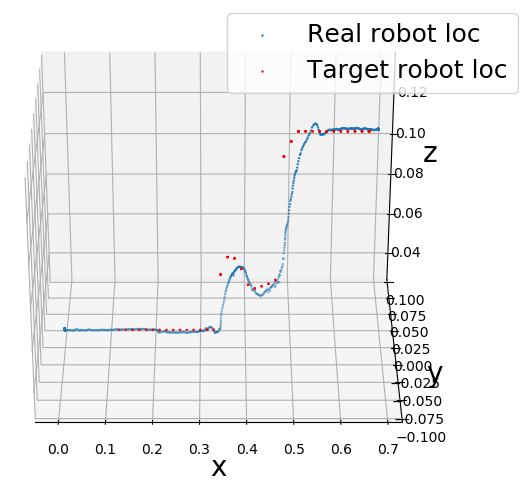}	
		\includegraphics[width=\locSize\linewidth]{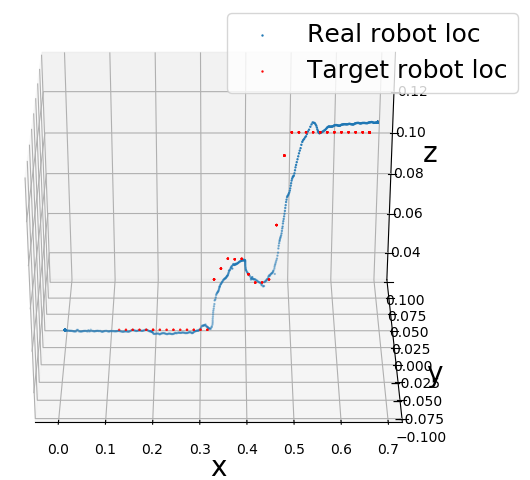}
		\includegraphics[width=\locSize\linewidth]{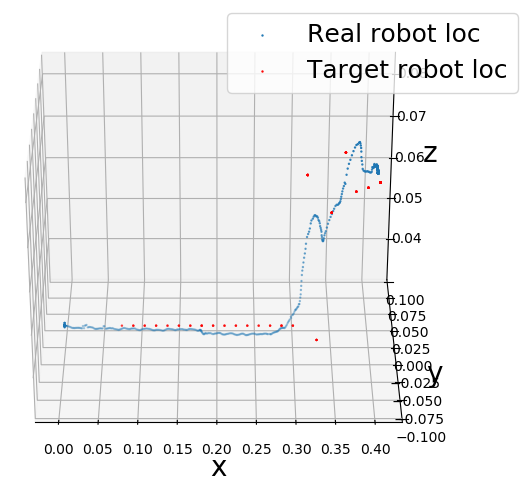}	
	}\\
	\subfloat[ramp]{
		\label{fig:simple_bias_loc:ramp}
		\includegraphics[width=\locSize\linewidth]{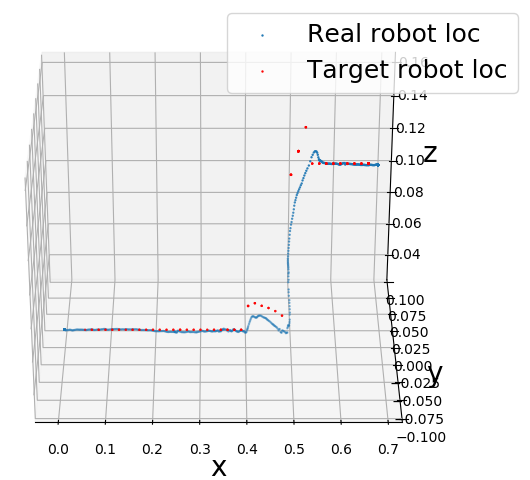}
		\includegraphics[width=\locSize\linewidth]{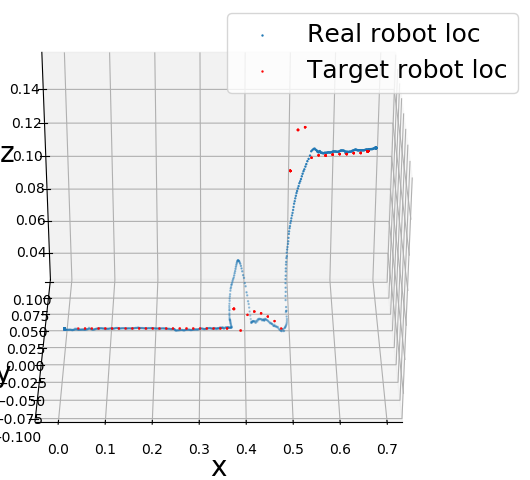}
		\includegraphics[width=\locSize\linewidth]{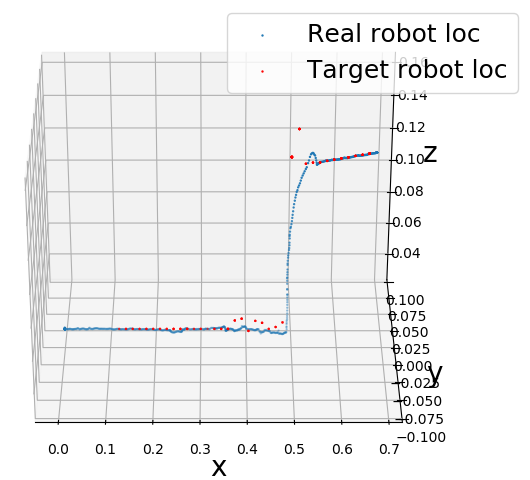}
		\includegraphics[width=\locSize\linewidth]{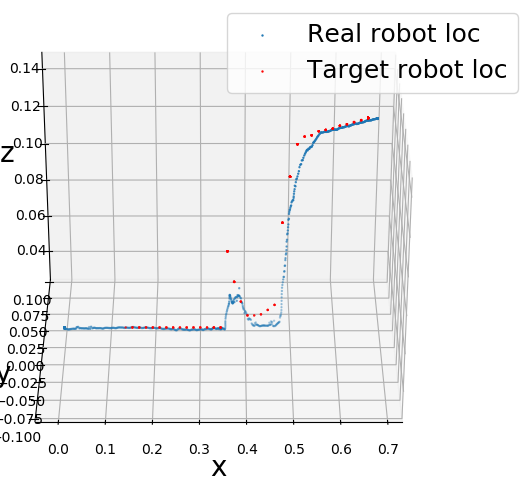}
		\includegraphics[width=\locSize\linewidth]{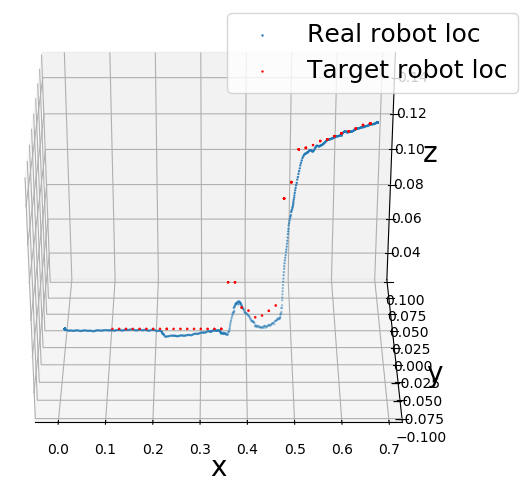}
		\includegraphics[width=\locSize\linewidth]{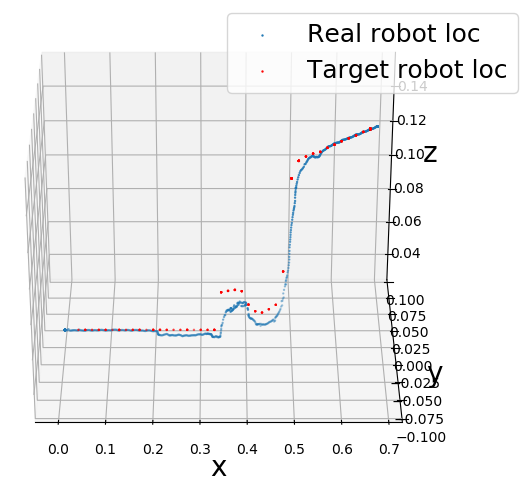}	
		\includegraphics[width=\locSize\linewidth]{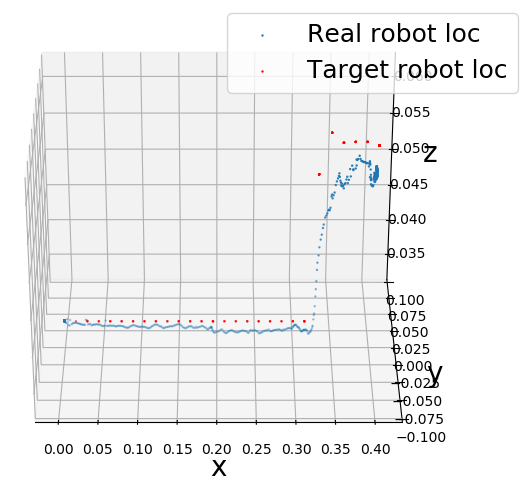}	
	}\\
	\subfloat[iramp]{
		\label{fig:simple_bias_loc:iramp}
		\includegraphics[width=\locSize\linewidth]{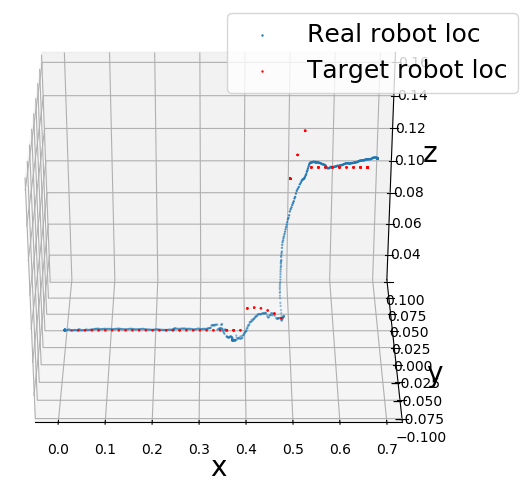}
		\includegraphics[width=\locSize\linewidth]{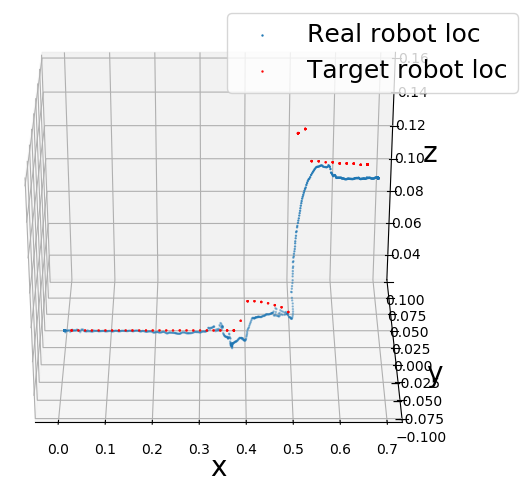}
		\includegraphics[width=\locSize\linewidth]{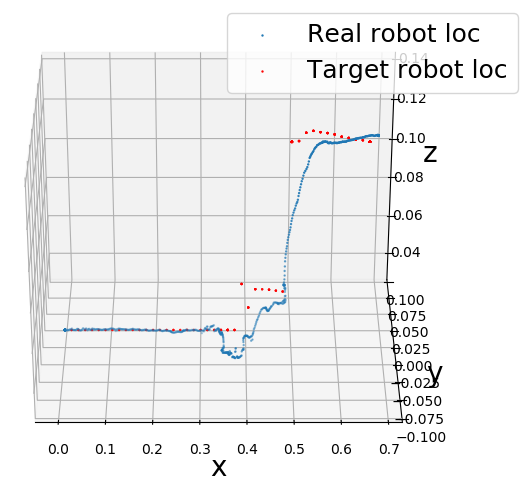}
		\includegraphics[width=\locSize\linewidth]{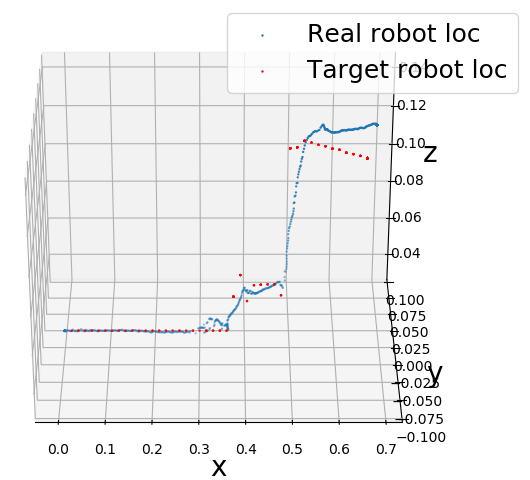}
		\includegraphics[width=\locSize\linewidth]{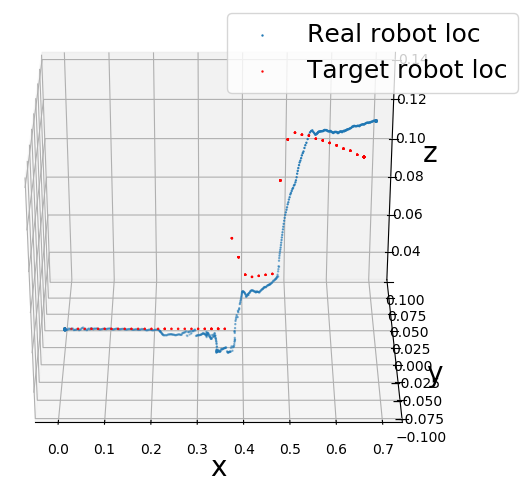}
		\includegraphics[width=\locSize\linewidth]{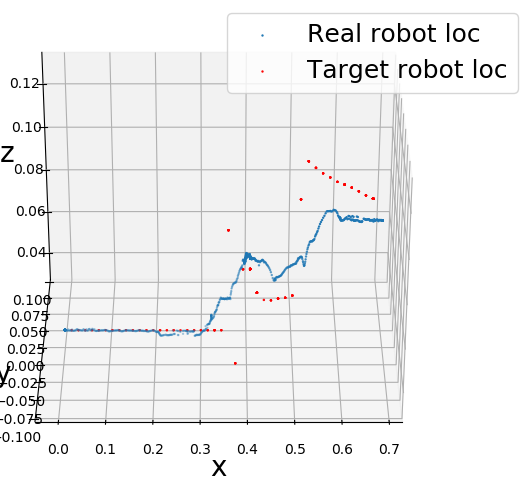}	
		\includegraphics[width=\locSize\linewidth]{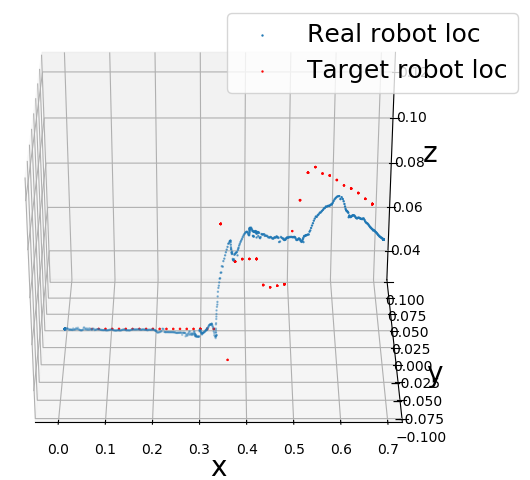}
		\includegraphics[width=\locSize\linewidth]{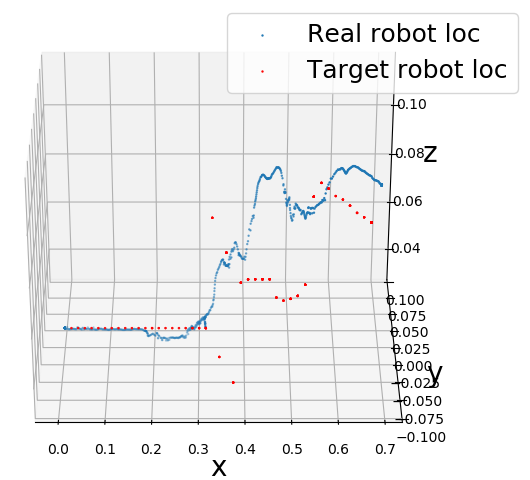}	
	}
	\caption{Position bias for terrains. The red is the path to follow, the blue points are real time location.}
	\label{fig:simple_bias_loc}
\end{figure*}

\section{Experiment}
\label{sec:exp}
\vspace{-0.2cm}
\subsection{Setting}

We apply our proposed method to our MARS Lab (Mobile Autonomous Robotic Systems Lab) rescue robot shown in Fig. \ref{fig:robot} and test if it can get over various terrains. 
Our small rescue robot has a wheel radius $r=0.035m$, track width $w_{track}=0.03m$ and robot width $w_{robot}=0.15m$. We use two DYNAMIXEL XM430-W210-T motors as wheel drive and four XM430-W350-T motors for the flippers. An Opti-track\footnote{\url{https://optitrack.com}} system is used to provide robot position and orientation. We don't use a mapping algorithm but simply provide a user-generated elevation map.

The algorithm is implemented in python. ROS\footnote{\url{https://www.ros.org}} is utilized for communication. 
Pose data from Opti-track is used with the VRPN server and the ROS client package\footnote{\url{http://wiki.ros.org/vrpn_client_ros}}.  Our program listens the signal from Opti-track, so the execution frequency is $100Hz$.

To standardize the experiment, we fix the robot initial pose and set the origin of the coordinate system as robot's initial location, robot's front direction as $x$-axis and its bottom-up as $z$-axis. Following right hand coordinate rule to uniquely determine the $y$-axis.

The test consists of three groups of test cases: step, ramp, inverse ramp,  all with various angles on the rotation axis (front center axis parallels to $z$-axis) that the robot straightly confronts to, as shown in Fig. \ref{fig:obstacle}. We call them (a) step, (b) ramp and (c) iramp, respectively. Since we use the clock-wise rotation, the difference between (b) ramp and (c) iramp is that the robot will confront the lower side or higher side of slope first.

The distance from the rotation axis to  $S_{middle, 2}$ is fixed as $0.54m$, with a rotation of the obstacle from $0^{\circ}$ to $40^{\circ}$, with a $5^{\circ}$ interval. (For $0^{\circ}$, robot faces the front surface of obstacle.)

We use those three cases plus their rotated variances for testing because on the front direction, they cover the scenes with (1) same left and right sides, (2) different left and right sides, (3) same distance from left and right sides to robot, (4) different distance from left and right sides to robot, (5) obstacle from low to high, (6) obstacle from high to low.


It should be noted that the robot is running autonomously in the experiments.

To get the robot location in the map, we pre-given the perfect 2.5D map of the obstacle and utilize tracking system (Opti-track) to provide the robot pose in real time. 
We also use the the Opti-track pose to easily evaluate the offset in the experiments. 

In the following we use our model on a real robot to cross real obstacles. Then we compute the error between the located pose and expected target pose.

\subsection{Real Robot Experiments with Various Obstacles}
\label{sec::exp::val}

\begin{table}[b!]
	\caption{Table Success Run. 1 is success, 0 is fail, - is not test.}
	\begin{tabular}{l| llllllll}
		\label{tab::success}
		& $0^\circ$ & $5^\circ$ & $10^\circ$ & $15^\circ$ & $20^\circ$ & $25^\circ$ & $30^\circ$ & $35^\circ$ \\ \hline
		step	& 1 & 1 & 1 & 1 & 1 & 1 & 1 & 0 \\
		ramp	& 1 & 1 & 1 & 1 & 1 & 1 & 0 & - \\
		iramp	& 1 & 1 & 1 & 1 & 1 & 0 & 0 & 0
	\end{tabular}
\end{table}

The attached video illustrates how the rescue robot gets over these three terrains with a range of rotation angles. 

\cite{yuan2019configuration} also tested the rotated step, but the planner was not aware of this. Here, in contrast, we planned with the elevation map for the complicated terrain. 
Table \ref{tab::success} shows that our robot can successfully move across the rotated cases. 

For step cases, the robot finished the $0^{\circ}-30^{\circ}$. However, when the rotation angle gets even larger, it fails the test. The video shows that the right front flipper gets stuck and Section \ref{sec::exp::bias} will try to reveal the according details in the data.

For ramp cases, when there is a rotation, the robot will confront the lower side of ramp first. And similar to 
step fail case, the right front flipper may get stuck.

The iramp cases are much harder than ramp, because the robot confronts the higher side of the ramp first when there's a rotation of ramp. A slip to the lower side commonly happens from the video record, though it climbs onto the ramp successfully, but we consider $25^{\circ}-35^{\circ}$ failed because its final configuration varies too much from the target. The stuck problem does not happen in such cases. 


\subsection{Configuration Error between Real Robot and Target}
\label{sec::exp::bias}



%

\addtolength{\belowcaptionskip}{-0.4cm}

\newcommand\blaSize{0.32}
\begin{figure}[tb]
	\centering
	\includegraphics[width=\blaSize\linewidth]{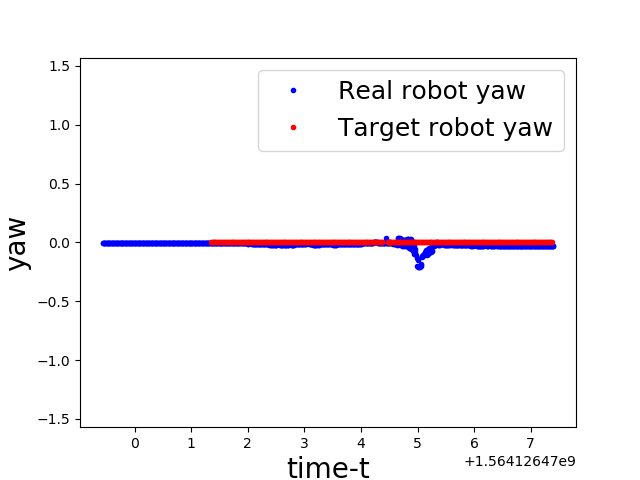}
	\includegraphics[width=\blaSize\linewidth]{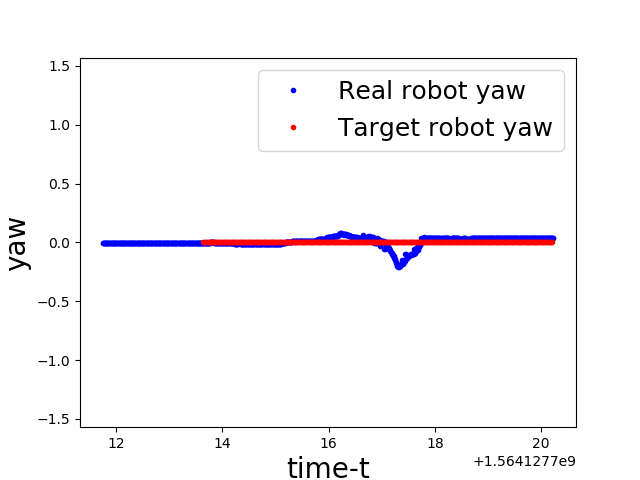}
	\includegraphics[width=\blaSize\linewidth]{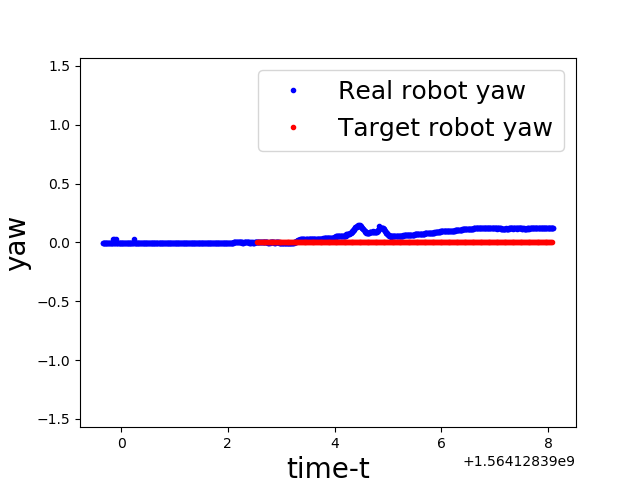}\\
	
	\includegraphics[width=\blaSize\linewidth]{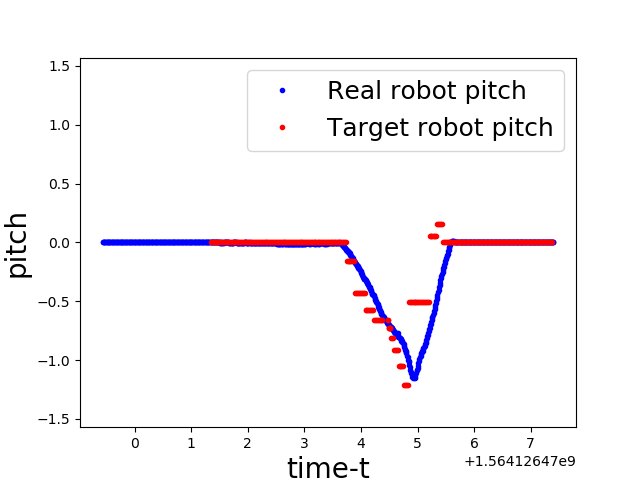}
	\includegraphics[width=\blaSize\linewidth]{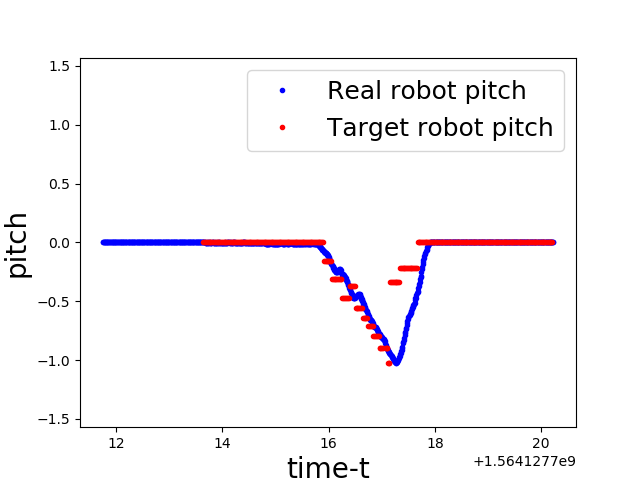}	
	\includegraphics[width=\blaSize\linewidth]{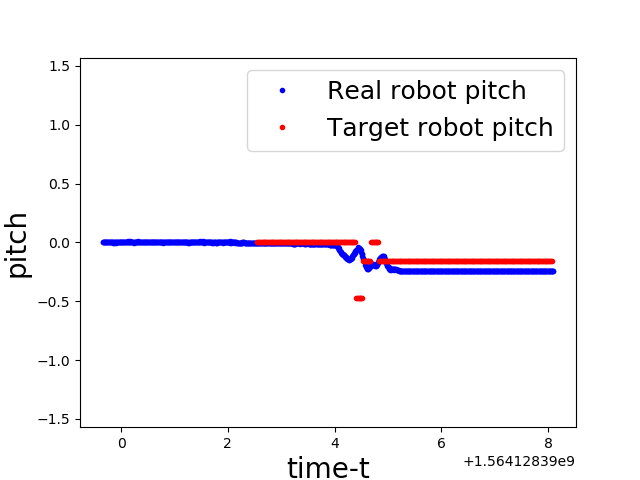}	\\

	\includegraphics[width=\blaSize\linewidth]{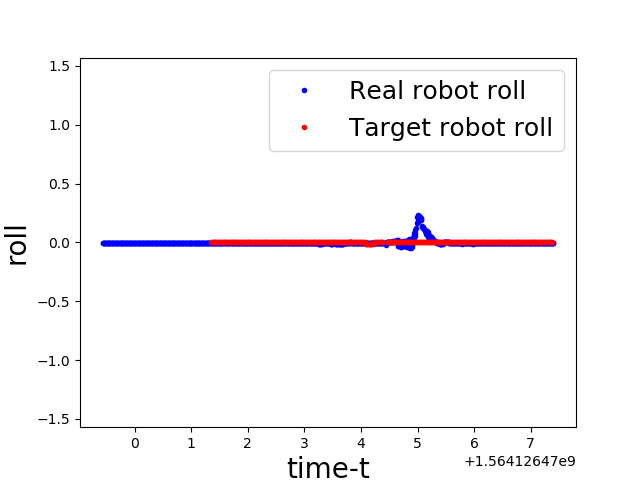}
	\includegraphics[width=\blaSize\linewidth]{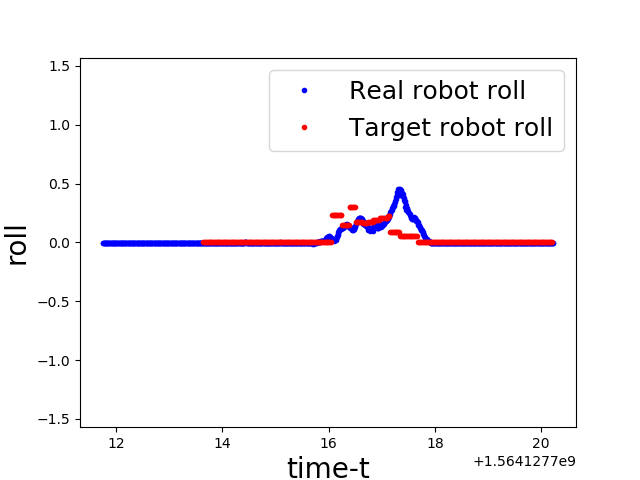}
	\includegraphics[width=\blaSize\linewidth]{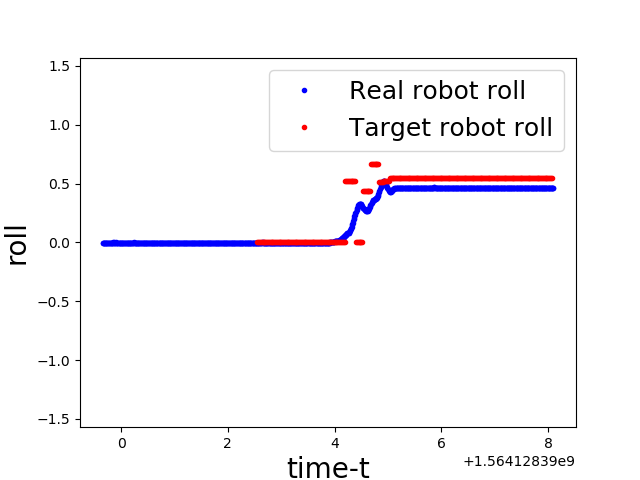}

	\caption{Orientation error for terrain step. Selected cases: $0^\circ$, $15^\circ$, $35^\circ$}
	\label{fig:step_ang}
\end{figure}
\begin{figure}[tb]
	\centering
	\includegraphics[width=\blaSize\linewidth]{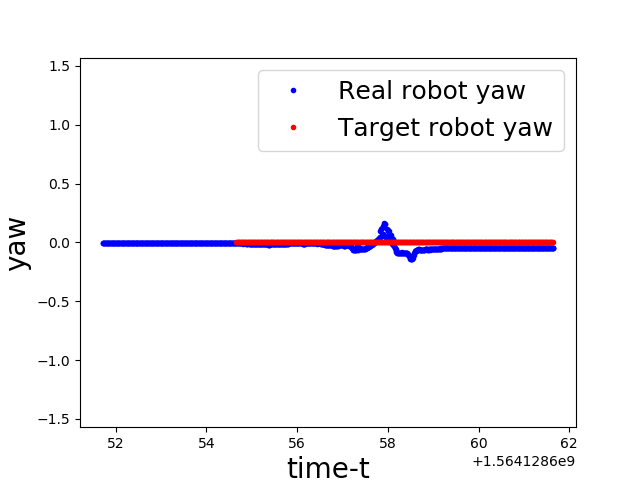}
	\includegraphics[width=\blaSize\linewidth]{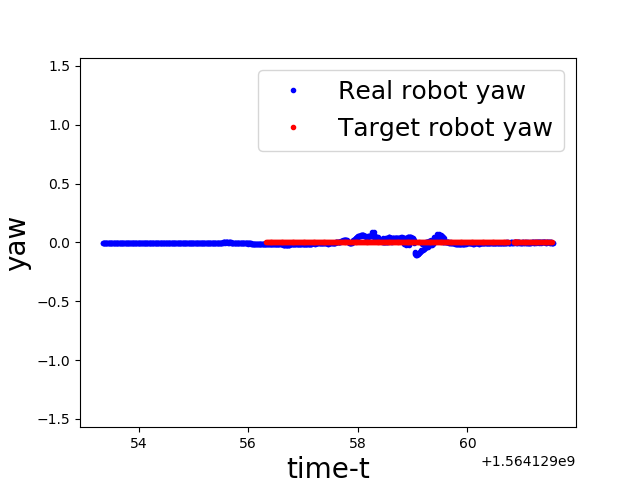}
	\includegraphics[width=\blaSize\linewidth]{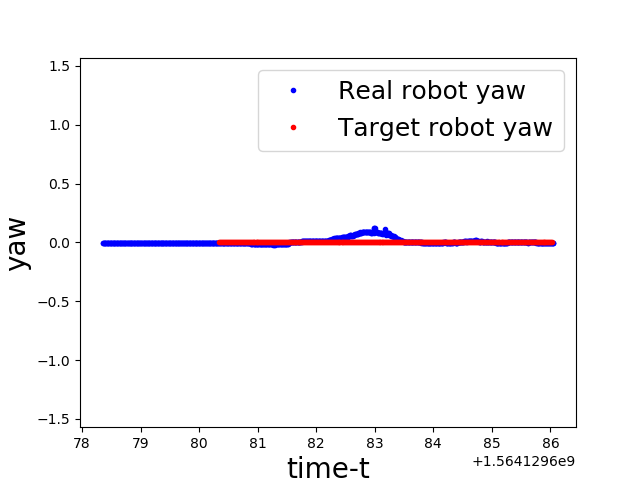}\\
	
	\includegraphics[width=\blaSize\linewidth]{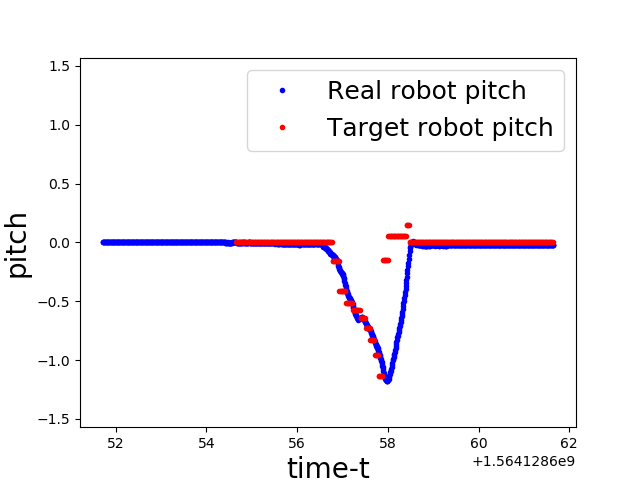}
	\includegraphics[width=\blaSize\linewidth]{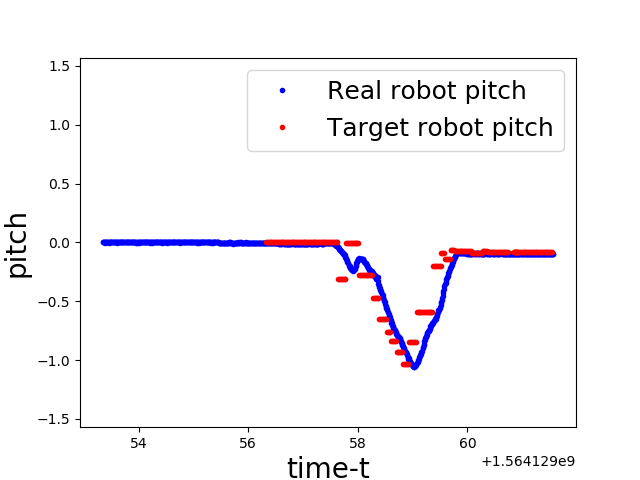}	
	\includegraphics[width=\blaSize\linewidth]{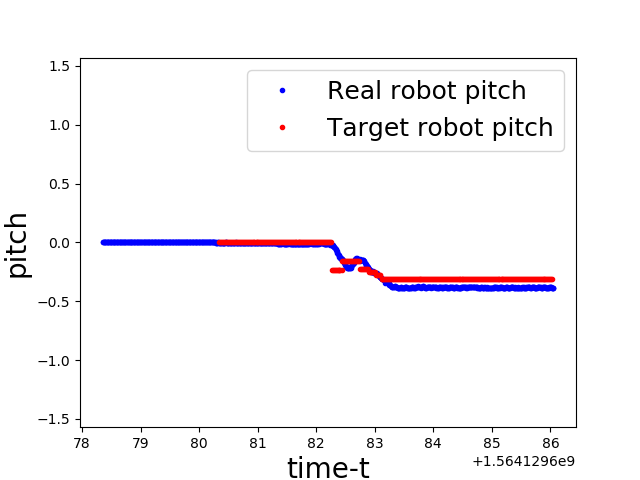}\\

	\includegraphics[width=\blaSize\linewidth]{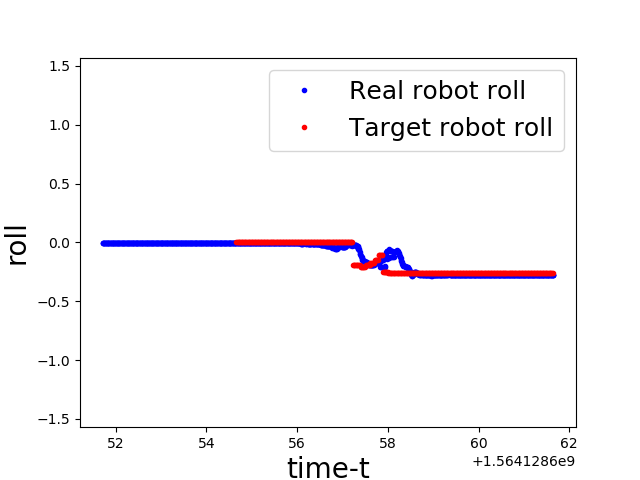}
	\includegraphics[width=\blaSize\linewidth]{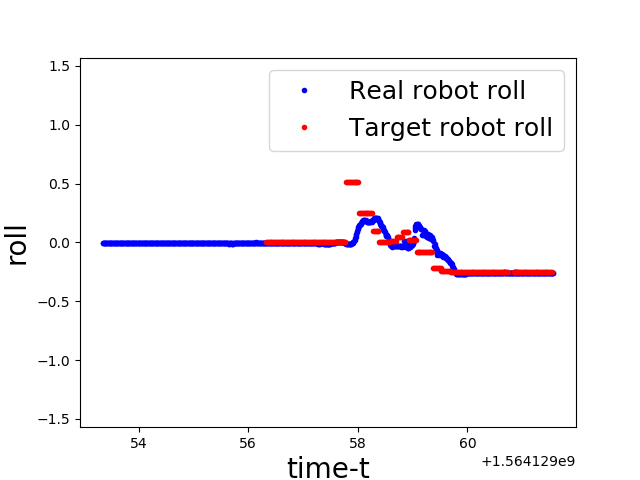}
	\includegraphics[width=\blaSize\linewidth]{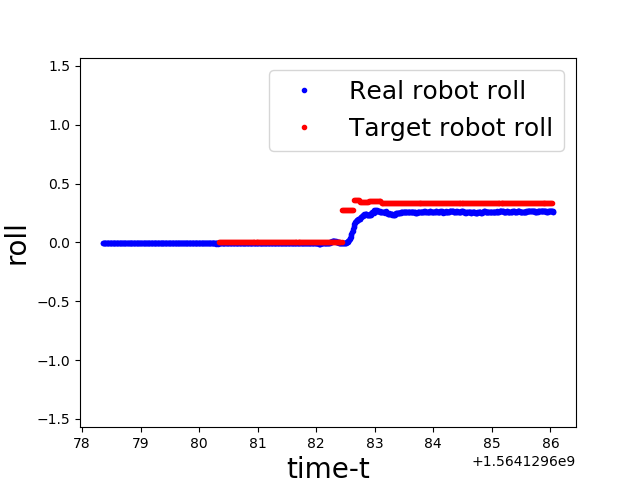}
	
	\caption{Orientation error for terrain ramp. Selected cases: $0^\circ$, $15^\circ$, $30^\circ$}
	\label{fig:ramp_ang}
\end{figure}
\begin{figure}[tb]
	\centering
	\includegraphics[width=\blaSize\linewidth]{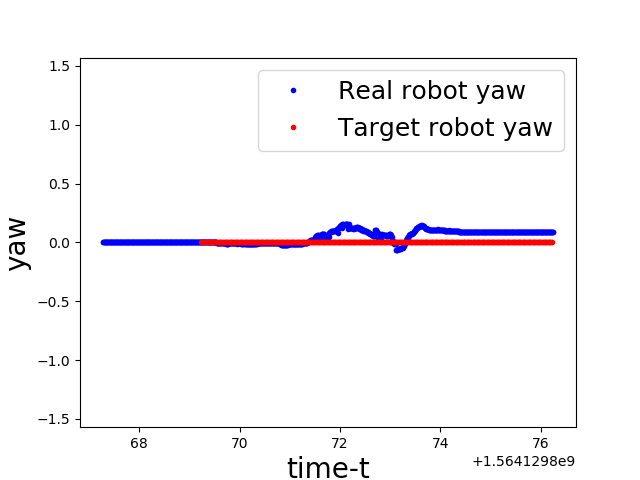}
	\includegraphics[width=\blaSize\linewidth]{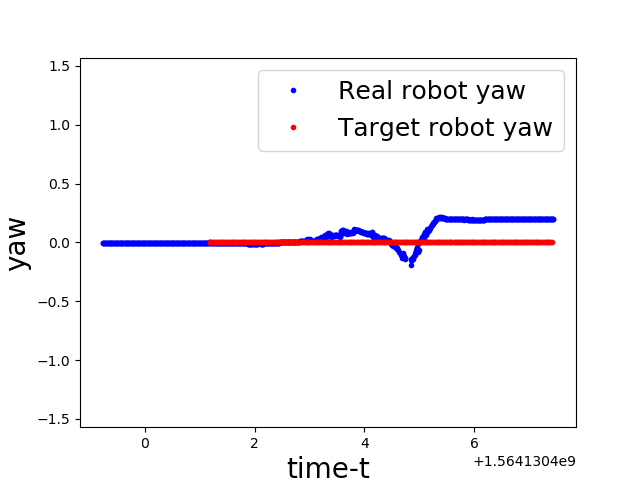}
	\includegraphics[width=\blaSize\linewidth]{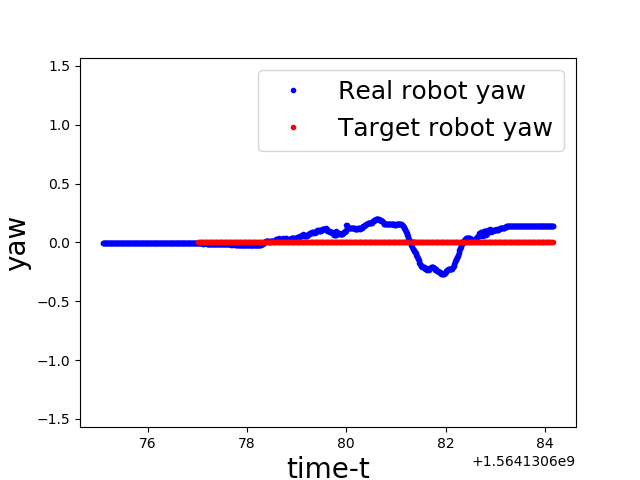} \\

	\includegraphics[width=\blaSize\linewidth]{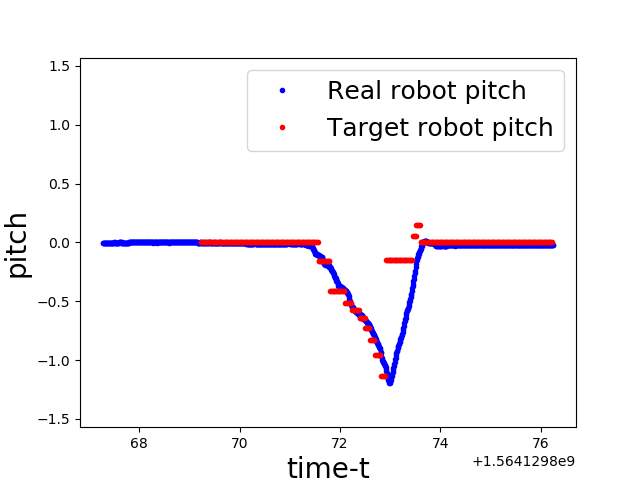}
	\includegraphics[width=\blaSize\linewidth]{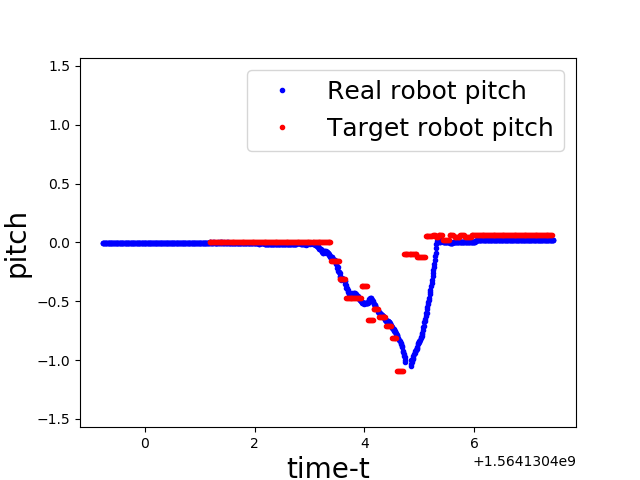}	
	\includegraphics[width=\blaSize\linewidth]{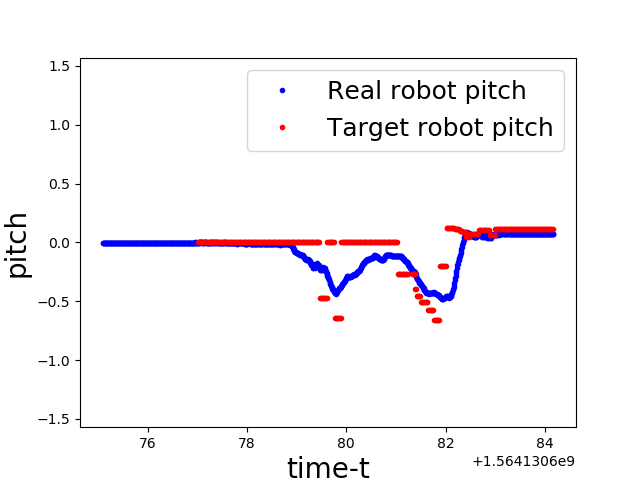} \\

	\includegraphics[width=\blaSize\linewidth]{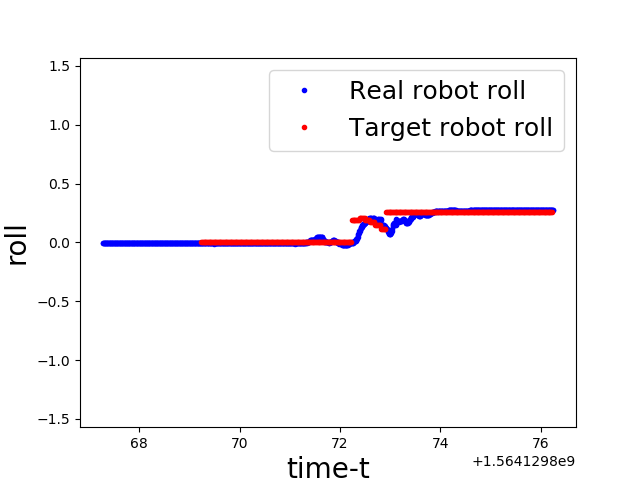}
	\includegraphics[width=\blaSize\linewidth]{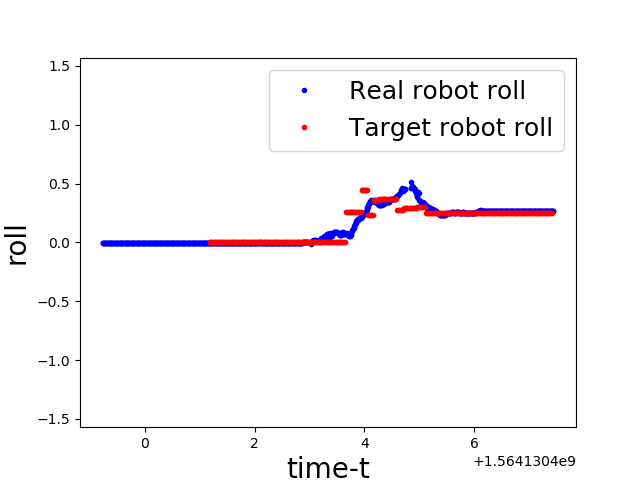}
	\includegraphics[width=\blaSize\linewidth]{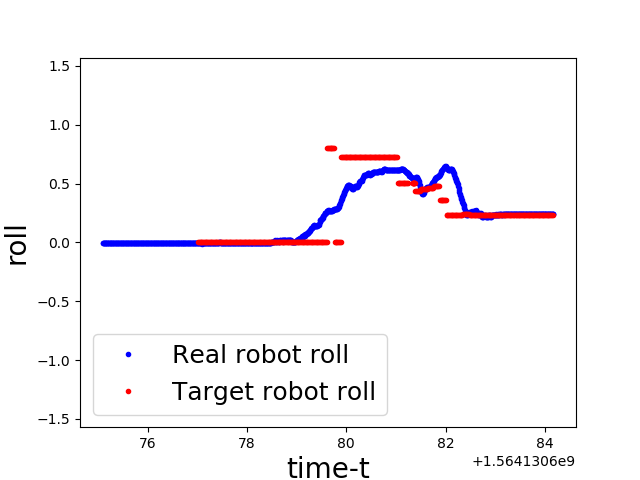} \\

	\caption{Orientation error for terrain iramp. Selected cases: $0^\circ$, $15^\circ$, $25^\circ$}
	\label{fig:iramp_ang}
\end{figure}
In this section, we ignore the record of the flipper angles, because our robot can provide accurate position of the flippers. For demonstration, only selected orientation errors of obstacles ($0^\circ$, $15^\circ$ and the first fail case) are shown in Fig. \ref{fig:step_ang}, \ref{fig:ramp_ang} and \ref{fig:iramp_ang}.

From Fig. \ref{fig:simple_bias_loc}, we can find that the $S_{middle,2}$ of succeed cases are well following the desire planned path. In Fig.~\ref{fig:simple_bias_loc} \ref{fig:step_ang}, \ref{fig:ramp_ang} and \ref{fig:iramp_ang}, the target parameters are discrete because robot updates the target of next step only when it reaches the current target.

In step with rotate angle $0^{\circ}-30^{\circ}$, Fig. \ref{fig:simple_bias_loc:step} and Fig. \ref{fig:step_ang} demonstrate the well traced morphology (location, orientation and flipper angle). However, when the rotation is as large as $35^{\circ}$, from video, we can find that the robot gets stuck with a flipper, blocking the movement.

In ramp with rotate angle $0^{\circ}-25^{\circ}$, Fig. \ref{fig:simple_bias_loc:ramp} and Fig. \ref{fig:ramp_ang} shows the performance of well tracing the configuration path. In such cases, when there is a rotation, the robot first confronts the lower side of the ramp. Similarly, when angle is large as $30^{\circ}$, its flipper stuck the movement as in video.

In iramp, the robot faces the higher side first when there is a rotation of the object. From Fig. \ref{fig:simple_bias_loc:iramp} and Fig. \ref{fig:iramp_ang}, when rotation is large, the robot slips to the lower side a little. When it is in $0^{\circ}-20^{\circ}$, robot still can well follow the configuration path. However, when it is larger, with the robot slips a lot, and the robot moves to a tremendously different setting, even though it climbs on the obstacle, as can be seen in the video.


\subsection{Discussion}
\vspace{-0.1cm}

From the recorded of positions and orientations from the tracking system, we find that the robot can follow the path very well, which demonstrates the applicability of our robot simplification and its flipper planning on the transformed representation.

However, there are some inaccuracy issues from the simplification and skeleton representation.

First, the modeling of the collision surface with the skeleton introduces an inaccuracy, because the collision between the skeleton and the inflated map is only equivalent to the touch of the simplified model and map, which is sometimes not specifically a collision between the real robot and map. 

Then, when there is a pitch, the real robot does not simply rotate around axis $S_{l,2}S_{r,2}$, as in the skeleton model. For example, without moving the track, if the robot rises its front side body up on the flat ground to change the pitch with $-\vartheta$, $S_{2}$ will move back $\vartheta R$. It is not a problem, since we can get the pose from Opti-track in real time and then compensate for it. However, similarly, when there is a roll, there will be a similar error on $y$-axis, which is not possible to make up. Thus this model is not as accurate for tracking the $y$-axis as it is for the $x$-axis. We consider this by carefully setting the cost on roll, which we can find in Section \ref{sec::exp::bias}, so the effect is very small.

\vspace{-0.2cm}
\section{Conclusions}
\vspace{-0.1cm}
\label{sec:conclusion}
We present an autonomous flipper planning method based on an elevation map and the robot model. We create a skeleton of the robot model and in turn inflate the 2.5D elevation map, to maintain correct collision checks. The planning algorithm manipulates the four flippers individually to traverse the 3D terrain. We implement the algorithm on a real robot and performed several experiments, showing that our rescue robot can get over various terrain with the proposed method. 

In the future we will improve the method by solving problems such as getting stuck. To allow general applicability of the method, Simultaneous Localization and Mapping (SLAM) should be introduced into this work for providing the elevation map and localization while following the planned path. Finally, we also plan to integrate the flipper planner into a global path planner.

\IEEEtriggeratref{17}
\bibliographystyle{IEEEtran}
\bibliography{references}
\end{document}